\documentclass[journal]{IEEEtran} 
\usepackage{times}
\usepackage{epsfig}
\usepackage{graphicx}
\usepackage{caption}
\usepackage{subcaption}
\usepackage{amsmath, mathtools}
\usepackage{amssymb}
\usepackage{authblk}
\usepackage{cancel}
\usepackage{mdframed}
\usepackage{booktabs}
\usepackage{comment}
\usepackage{xcolor}

\newcommand{\ignore}[1]{}

\newcommand{\ba}{\begin{array}}
\newcommand{\ea}{\end{array}}
\newcommand{\bc}{\begin{center}}
\newcommand{\ec}{\end{center}}
\newcommand{\be}{\begin{enumerate}}
\newcommand{\ee}{\end{enumerate}}
\newcommand{\bea}{\begin{eqnarray}}
\newcommand{\eea}{\end{eqnarray}}
\newcommand{\beas}{\begin{eqnarray*}}
\newcommand{\eeas}{\end{eqnarray*}}
\newcommand{\beq}{\begin{equation}}
\newcommand{\eeq}{\end{equation}}
\newcommand{\bfig}{\begin{figure}}
\newcommand{\efig}{\end{figure}}
\newcommand{\bi}{\begin{itemize}}
\newcommand{\ei}{\end{itemize}}
\newcommand{\bpic}{\begin{picture}}
\newcommand{\epic}{\end{picture}}
\newcommand{\btabular}{\begin{tabular}}
\newcommand{\etabular}{\end{tabular}}
\newcommand{\btable}{\begin{table}}
\newcommand{\etable}{\end{table}}

\newcommand{\es}{\vfill
                 \rule[-6mm]{170mm}{0.7mm} \\
                 \redw{{\tiny
		  \hfill S-\theslide}}
                 \end{slide}}

\newcommand{\matxx}[1]{{\mathtt #1}}
\newcommand{\vecXX}[1]{{\mathbf {#1}}}
\newcommand{\vecYY}[1]{{\boldsymbol {#1}}}

\providecommand{\etal}{{\em et~al.}}

\def \dbar {{\bar{d}}}

\def \hbar {{\bar{h}}}

\def \vecb {{\vecXX{b}}}

\def \vece {{\vecXX{e}}}

\def \vech {{\vecXX{h}}}

\def \vecm {{\vecXX{m}}}

\def \vect {{\vecXX{t}}}

\def \vecv {{\vecXX{v}}}

\def \vecx {{\vecXX{x}}}

\def \vecz {{\vecXX{z}}}

\def \veceta   {{\vecYY{\eta}}}
\def \vecmu    {{\vecYY{\mu}}}

\def \Xbar {{\bar{X}}}
\def \Ybar {{\bar{Y}}}

\def \matA {{\matxx{A}}}

\def \matI {{\matxx{I}}}
\def \matJ {{\matxx{J}}}

\def \matR {{\matxx{R}}}
\def \matS {{\matxx{S}}}

\def \matSigma  {{\matxx{\Sigma}}}
\def \matLambda {{\matxx{\Lambda}}}

\def \deg {^{\circ}}

\newcommand{\vecthree}[3]{\left(\begin{array}{c}#1\\#2\\#3\end{array}\right)}

\newcommand{\vectwo}[2]{\left(\begin{array}{c}#1\\#2\end{array}\right)}

\newcommand{\rowthree}[3]{\left(\begin{array}{ccc}#1&#2&#3\end{array}\right)}

\newcommand{\matthree}[9]{\left[\begin{array}{ccc}#1&#2&#3\\#4&#5&#6\\#7&#8&#9\end{array}\right]}

\makeatletter
\renewcommand*\env@matrix[1][*\c@MaxMatrixCols c]{%
  \hskip -\arraycolsep
  \let\@ifnextchar\new@ifnextchar
  \array{#1}}
\makeatother

\newcommand{\RR}{\mathbb{R}}

\newcommand{\SE}[1]{\ensuremath{\mathbf{SE}(#1)}}

\newcommand{\SO}[1]{\ensuremath{\mathbf{SO}(#1)}}

\newcommand{\se}[1]{\ensuremath{{\mathfrak{se}(#1)}}}

\usepackage[pagebackref=true,breaklinks=true,colorlinks,bookmarks=false]{hyperref}

\DeclareMathOperator\Exp{Exp}
\def \vectau    {{\vecYY{\tau}}}

\usepackage[normalem]{ulem}
\newcommand\redout{\bgroup\markoverwith{\textcolor{red}{\rule[.5ex]{2pt}{0.4pt}}}\ULon}

\newcommand\Jac[2]{\mathbf{J}^{#1}_{#2}}

\font\Bigmath=cmsy10 scaled \magstep2
\def\dplus{\mathrel{%
  \ooalign{$+$\cr\hss\lower.255ex\hbox{\Bigmath\char5}\hss}}}  
\def\dminus{\mathrel{%
  \ooalign{$-$\cr\hss\lower.255ex\hbox{\Bigmath\char5}\hss}}}

\begin{document}
\title{A Robot Web for Distributed Many-Device Localisation}
\author{Riku Murai, Joseph Ortiz, Sajad Saeedi, Paul H.J. Kelly, Andrew J. Davison
\thanks{
Riku Murai, Joseph Ortiz, Paul H.J. Kelly, Andrew J. Davison are with the Department of Computing, Imperial College London, UK (email: \{riku.murai15, j.ortiz, p.kelly, a.davison\}@imperial.ac.uk)
    
Sajad Saeedi is with the Department of Mechanical and Industrial Engineering, Toronto Metropolitan University, Toronto, Canada (email: s.saeedi@torontomu.ca)
}}


\maketitle

\begin{abstract}
We show that a distributed network of robots or other devices which make measurements of each other can collaborate to globally localise via efficient ad-hoc peer-to-peer communication. Our Robot Web solution is based on Gaussian Belief Propagation on the fundamental non-linear factor graph describing the probabilistic structure of all of the observations robots make internally or of each other, and is flexible for any type of robot, motion or sensor. We define a simple and efficient communication protocol which can be implemented by the publishing and reading of web pages or other asynchronous communication technologies.
We show in simulations with up to 1000 robots interacting in arbitrary patterns that
our solution convergently achieves global accuracy as accurate as a centralised non-linear factor graph solver while operating with high distributed efficiency of computation and communication. 
Via the use of robust factors in GBP, our method is tolerant to a high percentage of faulty sensor measurements or dropped communication packets.
Furthermore, we showcase that the system operates on real robots with limited onboard computational resources.
\end{abstract}

\section{Introduction}
\begin{figure}[t]
\centerline{
\hfill
\includegraphics[width=0.95\linewidth]{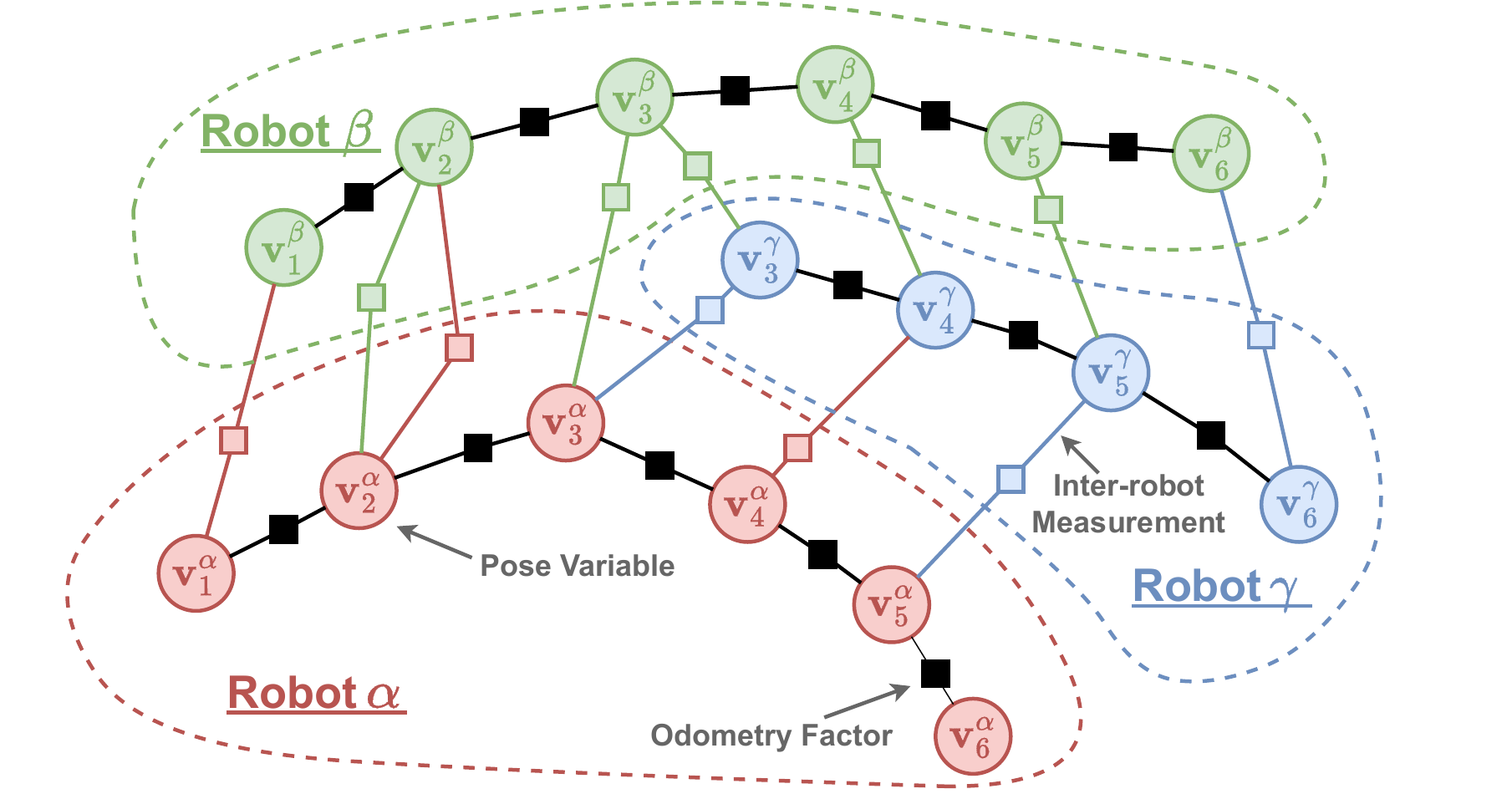}
\hfill
}
\caption{\label{fig:factor}
  In the Robot Web, we assume that a set of robots move through space while using their sensors to observe each other.
  The circles represent the variables -- where $\vecv_t^\alpha$ denotes a variable at timestamp $t$ which belongs to robot $\alpha$ --, and the squares are the factors. Robot $\gamma$ starts at timestamp 3 for clarity of visualisation.
  The full-factor graph for multi-robot localisation is used. Responsibility for storing and updating it is divided up between the multiple robots participating, as shown by the coloured regions separated by dotted lines.
  Each robot maintains its own pose variable nodes, odometry factors, and factors for the inter-robot measurements made by its sensors, and carries out continuous GBP on this graph fragment. Message passing across dotted line boundaries happens on an asynchronous and ad-hoc basis.
}
\vspace{2mm} \hrule
\end{figure}

As we head towards a future where embodied artificial intelligence is
ubiquitous, we expect that multiple robots, vehicles and other devices which share
the same environment will need to communicate and coordinate their
actions, whether their goal is explicit cooperation or just safe
independent action. One clear possibility is that all devices could
use a unified cloud-based `maps' system, presumably owned by one
company or government, which tracks and coordinates all devices. An
alternative, which we investigate here, is a distributed system-based
on per-device local computation and storage, and peer-to-peer communication between heterogeneous devices from different makers using standardised open protocols.
Inspired by the original
design of the World Wide Web, we call this concept the {\bf Robot Web}.

A key outstanding problem in multi-robot systems has
been true distributed localisation: how can a set of moving devices which move and observe each other within a space estimate their locations, using noisy actuators, sensors and realistic peer-to-peer communication?

In this paper, we present Robot Web, a solution to general, fully distributed and asynchronous many-robot localisation. Our solution is based
on the fundamental probabilistic factor graph
representation of perception and state estimation. We show that
Gaussian Belief Propagation (GBP) \cite{Ortiz:etal:ARXIV2021} is the key inference algorithm with
the appropriate properties of distributed processing/storage  and
message passing which
permits a convergent, exact solution to the full, dynamically changing
estimation problem via ad-hoc communication between robot peers.

In our solution, each robot stores and maintains its own part of the full factor graph (as shown in Figure~\ref{fig:factor}),
and updates and publishes a Robot Web Page of outgoing messages for
other robots to download and read whenever possible. Remarkably, using
GBP the whole factor graph can efficiently converge to localisation
estimates as accurate as full batch optimisation but without any
device ever needing to store or process more than its own local graph
fragment. Robots communicate via ad-hoc, asynchronous messages containing only small vectors and
matrices.  Significantly, GBP can deal with graphs
which have any type of parameterisation (e.g. 2D or 3D robot movement,
or any type of non-linear sensor measurements) and which can change dynamically
in arbitrary ways --- for instance, robots can join or
leave the web whenever needed, or reconfigure their sensors online.
We will show that it can also cope with and reject a high fraction of outlier measurements, for instance, caused by faulty sensors, and deal with highly unreliable communication channels.

Our approach is designed for  scalability. All communication is via a simple interface, and robots do not need any  privileged information about each other, such as even how many other robots are involved. The whole Robot Web therefore can be fully dynamic, with robots joining or leaving at will.
We believe that this formulation of many-robot localisation
could be the foundation for a new era of distributed Spatial AI. 

To summarise, the key contributions of our work are:
\begin{itemize}
    \item A distributed, scalable localisation system based on Gaussian Belief Propagation (GBP) --- {\bf Robot Web} --- that can localise thousands of robots in a simulated environment. Our method naturally handles asynchronous communication, communication failures, and a large number of outlier sensor measurements. Additionally, because all processing is local, our method can adapt to dynamic changes in the topology of the graph and handle disjoint connectivity.

    \item A framework for distributed multi-robot inference using GBP, and
    a novel extension of the GBP formulation to support Lie Groups, which is crucial for practical robotics applications.

    \item Extensive evaluation of our system to demonstrate the scalability. Using simulation, comparisons are made against a centralised counterpart across a range of parameters. Additionally, we have tested our system under challenging conditions, including the presence of a large number of non-Gaussian outlier sensor measurements and communication failures, in order to demonstrate its robustness.
    
    \item Real-time experiments with nine physical robots using only the limited computational resources available onboard. These experiments demonstrate the practicality and effectiveness of our approach and demonstrate that our method is feasible in real-world applications.
\end{itemize}

The rest of the paper is organised as follows. In Section~\ref{sec:related_works}, we review previous work on multi-robot localisation and distributed localisation in sensor networks. In Section~\ref{sec:gbp}, we provide an overview of GBP. 
Section~\ref{sec:lie_gbp} extends the formulation of GBP to support Lie Groups, which is necessary for practical robotics applications which require state estimation for both translation and rotation.
Section~\ref{sec:robotweb} describes our main contribution, which is a method for partitioning the factor graph across different devices and enabling communication between them. In Section~\ref{sec:demos}, we present experimental results in a simulated environment and in Section~\ref{sec:realworld} with real robots. In Section~\ref{sec:ongoing}, we outline ongoing research topics. Finally, in Section~\ref{sec:conclusion}, we conclude the paper and discuss potential directions for the future of distributed Spatial AI.

\section{Related Work}
\label{sec:related_works}
Robot Web uses the standard Gaussian factor graph representing the multi-robot localisation problem and is most closely related to the wealth of factor graph formulations and solvers in robotics, as well explained in the work of Dellaert and Kaess
\cite{Dellaert:Kaess:Foundations2017,Dellaert:AR2021}. 
Most methods for inference on factor graphs assume a centralised computer with access to the whole graph and focus on either efficient batch solution or incremental inference on graphs that are continually
changing.
Centralised pose-graph optimisation algorithms suitable for multiple robots are well-explored in the literature \cite{Indelman:etal:ICRA2014, Bailey:etal:ICRA2011, Kim:etal:ICRA2010, Andersson:etal:ICRA2008}.
MR-iSAM2 \cite{Yetong:etal:IROS2021} extends iSAM2~\cite{Kaess:etal:IJRR2012} to build an incremental, centralised graph optimisation method for multiple robots.
However, centralised methods require a base station, and are vulnerable to failure of this station, can require high communication bandwidth, create privacy concerns, and generally are not scalable~\cite{Lajoie:etal:ArXiv2022}.

There have been many previous attempts  at  distributed multi-robot
localisation and SLAM which uses local computation and
peer-to-peer communication, but these are generally far more limited
or specific than Robot Web; there is a useful recent survey by Halsted~\etal~\cite{Halsted:etal:ArXiv2021}.
Leung~{\it et~al.} demonstrated that in a distributed network of robots, where the network is never fully connected, it is possible to achieve exact and centralised-equivalent estimates for localisation \cite{Leung:etal:TRO2010} and SLAM \cite{Leung:etal:JIRS2015}. 
Many of the existing results in multi-robot SLAM perform a version of gradient descent, albeit in a distributed fashion~\cite{Murphey:Fan:IROS2020}. A recent example is the work by Tian {\it et al.}~\cite{Tian:etal:TRO2021} utilising block coordinate descent on the Riemannian manifold. The work by
Tian {\it et al.} has also been used in Kimera-Multi~\cite{Yun:etal:ICRA2021} for pose-graph optimisation.

Other papers formulate the graph optimisation problem as a linear minimum mean-squared-error problem and utilise stationary iterative methods \cite{Hageman:Young:1981} such as the Jacobi method or the Gauss-Seidel method.
For example, the Jacobi method is used by Barooah and Hespanha \cite{Barooah:Hespanha:ISIP2005} and Aragues {\it et al.} \cite{Aragues:etal:ICRA2011} to solve the linearised system equations iteratively. The Jacobi method is implemented in a distributed manner as described in \cite{Delouille:etal:IPSN2004}.
Choudhary {\it et al.} \cite{Choudhary:etal:IJRR2017} demonstrates a distributed algorithm that scales to 50 robots.
The algorithm utilises a distributed Gauss-Seidel method for solving
linearised equations. This method has been used as a back-end optimisation module in other works such as DOOR-SLAM \cite{Lajoie:etal:RAL2020} and the recent work by Cieslewski {\it et al.} \cite{Cieslewski:etal:ICRA2018}.

In DDF-SAM, Cunningham {\it et al.}~\cite{Cunningham:etal:IROS2010} presented an algorithm that distributes factor graph optimisation across multiple robots. It relies on Gaussian elimination, and requires robots to exchange Gaussian marginals about shared variables. 
Later in DDF-SAM2 \cite{Cunningham:etal:ICRA2013}, an extension is proposed that avoids information double-counting.
Similarly, the method presented in~\cite{Lazaro:etal:IROS2011} involves robots exchanging condensed portions of their factor graphs in order to minimise communication.
However, approaches which rely on Gaussian elimination become slow as the number of variables grows.

A significant limitation of the distributed methods above is that they
are synchronous, which means the robots must share their messages at
predetermined times to ensure the shared information is up-to-date.
In contrast, asynchronous methods offer the flexibility that the
robots can operate at their own rate, without waiting for other
robots. Examples are the works by Todescato {\it et al.}
\cite{Todescato:etal:ECC2015} and Tian {\it et al.}
\cite{Tian:etal:RAL2020}. The first reference conducts the
optimisation in Euclidean space, while the latter does the
optimisation by computing the gradient descent on a Riemannian
manifold in a distributed manner. The authors of these papers also
investigated the convergence of their distributed algorithms under
communication delay.

Robot Web is also asynchronous and distributed, but both much more
simple in formulation and more general than these methods. Significantly, Tian~{\it et~al.}~\cite{Tian:etal:RAL2020} assume Gaussian noise are unable to handle outliers.  Additionally, their formulation requires the sensor measurements to be a relative transformation.
Robot Web supports general robot and sensor models and allows robust
factors, making it robust to large fractions of non-Gaussian outlier measurements.

There has been some work on
 multi-agent distributed localisation using variations of belief
 propagation in the sensor networks community.
For instance, Schiff {\it et al.} \cite{Schiff:etal:IROS2009} performed multi-robot localisation using non-parametric belief propagation.
Wymeersch {\it et al.} \cite{Wymeersch:etal:PIEEE2009} also used belief propagation to perform cooperative positioning in a distributed manner. They used the sum-product algorithm over a factor graph in an ultra-wideband network.
Then Caceres {\it et al.} \cite{Caceres:etal:JSAC2011} extended \cite{Wymeersch:etal:PIEEE2009} to a network composed of GNSS nodes.
These distributed positioning methods were later generalised by adding
nonlinear measurement models and utilising Gaussian message passing
\cite{Bin:etal:WCSP2014}, \cite{Li:etal:SP2015}; and
in 2017, Wan {\it et al.} \cite{Wan:etal:IJDSN:2017} proposed a distributed multi-robot SLAM algorithm, using belief propagation. In their method, a mixture of Gaussian and non-parametric models was used to handle nonlinear models. They also assumed measurements are affected by Gaussian noise and used synchronous
message passing.

Robot Web goes far beyond these methods to present a general framework for general robots and sensors.
It defines for the first time an open, asynchronous communication framework,
and via the focus on GBP with robust factors enables highly robust and scalable performance.

\section{Gaussian Belief Propagation}
\label{sec:gbp}
A factor graph has a bipartite structure, composed of variable nodes and factor nodes. Variable nodes are only connected to factor nodes and vice versa.

GBP performs marginal inference, where it computes the per-variable marginal posterior distributions. The marginal inference proceeds iteratively via message passing between variable and factor nodes, which can in principle happen in many different types of patterns but still with convergent behaviour.

GBP is guaranteed to compute the exact marginal means on convergence, although the same is unfortunately not true for the variances as they are often overconfident for loopy graphs~\cite{Weiss:Freeman:NIPS1999}. Thus when it converges, GBP yields the same mean as the maximum a posteriori (MAP) inference produced by a centralised solver. 
There are generally no convergence guarantees for GBP, but there are certain conditions under which it is known to converge and methods that can improve its chances of convergence~\cite{Bickson:PhDThesis:2008}.

This section summarises the basic operation of GBP.
For more detailed derivation, refer to FutureMapping~2~\cite{Davison:Ortiz:ARXIV2019} or `A visual introduction to Gaussian Belief Propagation'~\cite{Ortiz:etal:ARXIV2021}.

We represent the Gaussian distribution in information form as:
\beq
    \mathcal{N}(\vecx; \vecmu, \matSigma) = \mathcal{N}^{-1}(\vecx; \veceta, \matLambda)~,
\eeq
where $\matLambda = \matSigma^{-1}$ and $\veceta = \matLambda \vecmu$.
In GBP where we have $N_v$ variables and $N_f$ factors, variables $V = \{\vecv_i\}_{i=1:N_v}$ are assumed to be Gaussian; thus, each variable has a belief  $b(\vecv_i) = \mathcal{N}^{-1}(\vecv_i; \veceta_i, \matLambda_i)$. 
Factors $F = \{f_i\}_{i=1:N_f}$ are a probabilistic Gaussian constraint on or between the variables.
We define neighbourhood $n(\vecx)$ to indicate the set of adjacent nodes connected via an edge to node $\vecx$ in the factor graph, where $\vecx$ may be a variable or a factor.

GBP can be divided into local operations at the variables, local operations at the factors, and message passing between them which we now detail.

\subsection{Variable Belief Update}
\label{sec:variable_belief_update}
The belief of a variable is the product of all the incoming messages:
\beq
b(\vecv_i) = \prod_{f\in n(\vecv_i)} \vecm_{f\rightarrow \vecv_i}(\vecv_i)~,\label{eq:variable_belief_update}
\eeq
where
$\vecm_{f \rightarrow \vecv_i}(\vecv_i) = \mathcal{N}^{-1}(\vecv_i; \veceta_{f \rightarrow \vecv_i}, \matLambda_{f \rightarrow \vecv_i})$ 
is a message from a factor to a variable $\vecv_i$ along an edge.

\subsection{Variable to Factor Message}

The message from a variable node $\vecv_i$ to a factor node $f_j \in n(\vecv_i)$ is computed as the product of all incoming messages from the neighbouring factor nodes of $\vecv_i$, except for the message from $f_j$. This can be expressed as:

\beq
\vecm_{\vecv_i \rightarrow f_j}(f_j) = \prod_{f \in n(\vecv_i) \backslash f_j} \vecm_{f \rightarrow \vecv_i}(\vecv_i)~.\label{eq.variable_to_factor_message}
\eeq

\subsection{Factor Potential of Linear and Non-linear Function}
The  general definition of a Gaussian factor is:
\bea
\label{equ:generalfactor}
f(\vecx; \vecz) &=& K e^{-\frac{1}{2} \left[ (\vecz - \vech(\vecx))^\top \matLambda_s (\vecz - \vech(\vecx))   \right]  } \\
         &=& K e^{-E(\vecx; \vecz)  }
~,
\eea
where we have defined the general factor energy: 
\beq
    E(\vecx; \vecz) = \frac{1}{2} (\vecz - \vech(\vecx))^\top \matLambda_s (\vecz - \vech(\vecx))
    ~.
    \label{equ:general_factor_energy}
\eeq

This expression represents the probability of obtaining vector measurement $\vecz$ from the sensor as a function of the set of involved variables $\vecx$.
Here, $\vech$ is the
functional form of the dependence of the measurement on state
variables. Matrix
$\matLambda_s$ is the precision of the
measurement. 
$\vecx$ represents the state space of
all of the $N$ variables connected to the factor.

The general formula in Equation~\eqref{equ:generalfactor} for a factor represents a Gaussian distribution over the observed measurement $\vecz$.
For linear measurement functions $\vech(\vecx)$, the resulting linear factor is also a Gaussian distribution in the variables $\vecx$.
To derive the Gaussian factor over $\vecx$, we begin with our general factor energy in Equation~\eqref{equ:general_factor_energy}, and our goal is to manipulate the energy into the form of a Gaussian distribution over $\vecx$ in the information form: 
\beq
    E(\vecx; \vecz) = \frac{1}{2} \vecx^\top \matLambda \vecx - \veceta^\top \vecx
    ~.
    \label{equ:energy_inf_form}
\eeq
After transforming the energy into this form, we can identify the parameters $\veceta$ and $\matLambda$ of the factor distribution.

To begin with, any linear measurement function can be generally written as: 
\begin{equation}
    \vech (\vecx) = \matA \vecx + \vecb
    ~,
    \label{equ:linear_meas_fn}
\end{equation}
where $\vecb \in \RR^m$ is a constant vector and $\matA \in \RR^{m \times n}$ is a constant matrix for $\vecz \in \RR^m$ and $\vecx \in \RR^n$.

Substituting Equation~\eqref{equ:linear_meas_fn} into the Equation~\eqref{equ:general_factor_energy} and rearranging:
\bea
    E(\vecx; \vecz) &=& \frac{1}{2} \left[ \vecz - \matA \vecx - \vecb \right]^\top \matLambda_s \left[\vecz - \matA \vecx - \vecb \right]  \nonumber \\
    &=& \frac{1}{2} \left[ (\vecz - \vecb) - \matA \vecx \right]^\top \matLambda_s \left[ (\vecz - \vecb) - \matA \vecx \right]  \nonumber \\
    &=&  \frac{1}{2} \bigg[ (\vecz - \vecb)^\top \matLambda_s (\vecz - \vecb)
    + (\matA \vecx)^\top \matLambda_s \matA \vecx   \nonumber \\
    && - (\vecz - \vecb)^\top \matLambda_s \matA \vecx  
    - (\matA \vecx)^\top \matLambda_s (\vecz - \vecb)  \bigg]
    ~.
\eea

The first of the four terms here is a constant which doesn't depend on $\vecx$ and so we can drop it into the normalising constant. The third and fourth are equal (one is the transpose of the other, and both are scalars), so we can simplify to:
\bea
    E(\vecx; \vecz) &=& \frac{1}{2} (\matA \vecx)^\top \matLambda_s \matA \vecx  - (\vecz - \vecb)^\top \matLambda_s \matA \vecx  \nonumber \\
    &=&  \frac{1}{2} \vecx^\top (\matA^\top  \matLambda_s \matA) \vecx  - \big( \matA^\top \matLambda_s (\vecz - \vecb) \big)^\top \vecx
  ~.
\eea

Matching this with the Equation~\eqref{equ:energy_inf_form}, we can identify the Gaussian information form parameters of the linear factor as:
\bea
\veceta_f &=& \matA^{\top} \matLambda_s(\vecz - \vecb), \label{equ:linear_factor_potential_eta}
\\
\matLambda_f &=& \matA^{\top} \matLambda_s \matA \label{equ:linear_factor_potential_Lam}~.
\eea

For a non-linear measurement function $\vech(\vecx)$, we can use a first-order Taylor expansion to linearize it around a reference point $\vecx_0$: $\vech(\vecx) \approx \vech(\vecx_0) + \matJ(\vecx - \vecx_0)$,
where $ \matJ$ is the Jacobian matrix of $\vech(\vecx)$ with respect to $\vecx$ evaluated at $\vecx_0$.
Rearranging and then matching, we get:
$\matA = \matJ$ and $\vecb = \vech(\vecx_0) - \matJ\vecx_0$.
Hence, the factor potential of a non-linear measurement function is:
\bea
\label{equ:non-linear-factor-potential}
\veceta_f &=& \matJ^{\top} \matLambda_s(\vecz - (\vech(\vecx_0) - \matJ\vecx_0))~, \\
\matLambda_f &=& \matJ^{\top} \matLambda_s \matJ~.
\eea

\subsection{Factor to Variable Message}
Let $V_f = n(f)$, a set of neighbouring variable nodes to factor $f$.
To compute factor to variable messages, we take the product of the factor potential and messages from $V_f$, except for $\vecv_i$ before marginalising out 
$V_f \backslash \vecv_i$:
\beq
\vecm_{f\rightarrow \vecv_i}(\vecv_i) = \sum_{V_f \backslash \vecv_i} f(V_f) \prod_{\vecv_j \in V_f \backslash \vecv_i} \vecm_{\vecv_j\rightarrow f}(\vecv_j)~.\label{eq.factor_to_variable_message}
\eeq

\subsection{Robust Factors}
A common misunderstanding among
roboticists is that loopy GBP is not a useful algorithm 
because uncertainty gating for data association is not possible due to overconfident covariance. 
A little consideration of any particular variable in the graph at
convergence explains why we can in fact do properly-formulated data
association using GBP. For a variable, we calculate its marginal
belief by multiplying the probability distributions from the latest
messages from all the factors it connects to. The fact that the marginal mean
is correct for all variables at convergence shows that the {\em
  relative} precisions of local messages into a variable are convergently
correct. {\em This} is what is important for data association --- to be able to judge multiple hypotheses represented by different factors in the graph against each other.

We use robust factors on inter-robot measurements, and
our approach can straightforwardly deal with a high
fraction of non-Gaussian outlier sensor measurements. 
These outlier measurements do not need to be identified or dealt with in any explicit manner, but are simply automatically down-weighted by the robust factors and end up having a negligible influence on the whole graph. Note that this is a kind of `lazy data association' because GBP's final commitment to whether to accept a measurement can often happen some time after the measurement is reported, or could even be reversed later on if other supporting measurements arrive.

Specifically, using all of the latest incoming messages at a factor, we can compute the current potential at that factor,  and interpret this as a Mahalanobis distance which indicates the `energy' of the factor in terms of how many standard deviations away from its most probable value the current states of the variables are. This can be used for visualisation (to see which factors in a network are currently the most `stretched' during optimisation), or for applying the effect of a  robust kernel to downweight its effect.

At a factor, we combine the latest states of all the connected variables to form a stacked state representation $\vecx_0$.
This acts as a linearisation point, and is used to compute the current Mahalanobis distance of the factor: 
\beq
M = \sqrt{(\vecz - \vech(\vecx_0))^{\top} \matLambda (\vecz - \vech(\vecx_0))}
\eeq

Robust measurement functions in GBP can be handled by applying
a robust scaling factor $k_R$  to the precision matrix and the
information vector of the factor potential~\cite{Davison:Ortiz:ARXIV2019}:
\bea
\tilde\matLambda' &=& k_R \matLambda' \\
\tilde\veceta_\mu &=& k_R \veceta_\mu
~.
\eea
This scaling factor depends on the Mahalanobis distance and can implement any robust kernel such as Huber, Tukey or DCS~\cite{Agarwal:etal:ICRA2013}.
Through this mechanism, a factor will weaken itself by the appropriate probabilistic amount when the combination of variables it connects to means that the measurement the factor represents is likely to be an outlier. Overall, when many robust factors are connected in a large factor graph, GBP is able to use this mechanism to achieve lazy, reversible data association. We will see the impact of this in our results later in Section~\ref{sec:outlier}, where we show that we are able to deal with a large proportion of incorrect measurements in multi-robot localisation.

\subsection{Gaussian Belief Propagation for Multi-Device Distributed Inference}
Extending GBP to the multi-device system is straightforward.
Since GBP is a node-wise distributed algorithm and it operates via message passing, centralised GBP and distributed GBP are identical under perfect communication. No changes to the core algorithm is required for distributed inference.

As shown in Figure~\ref{fig:factor}, we separate the nodes of the graph amongst the robots.
At each iteration, we perform the following steps:
\subsubsection{Message Computation} Locally on each of the robots, variable-to-factor messages are computed using Equation~\eqref{eq.variable_to_factor_message}, and the factor-to-variable messages are computed using Equation~\eqref{eq.factor_to_variable_message}. 
\subsubsection{Message Exchange} Messages are exchanged internally within a robot and externally between robots. 
An internal message is a message which is exchanged between nodes that belong to the same local graph, and an external message is a message which is sent from one robot to another via some communication mechanisms.
To explain the external message passing more concretely, we use Robot $\alpha$ and Robot $\beta$ from Figure~\ref{fig:factor} as an example. The factors of Robot $\alpha$ will send factor-to-variable messages to $\vecv_1^\beta$, $\vecv_2^\beta$, and variable-to-factor messages from $\vecv_2^\alpha$, $\vecv_3^\alpha$ together with its point-estimate $\mu_{\vecv_2^\alpha}$, $\mu_{\vecv_3^\alpha}$ as the linearisation point. Both factor-to-variable and variable-to-factor messages are a marginal distribution, which is a $N\times 1$ vector and a $N\times N$ precision matrix for the message to or from a variable with $N$ degrees of freedom.
Similarly, Robot $\beta$ will send Robot $\alpha$ the factor-to-variable messages to $\vecv_2^\alpha$, $\vecv_3^\alpha$, and variable-to-factor messages from $\vecv_1^\beta$, $\vecv_2^\beta$ with its point-estimate means.
\subsubsection{Variable Belief Update}
Once messages are exchanged, the beliefs of the variables are updated via Equation~\eqref{eq:variable_belief_update}, and we repeat this process until convergence.

We cover the specifics of how Robot Web partitions the factor graph, the communication model, and how the robots discover other robots' factors later in Section~\ref{sec:robotweb}.

\section{Gaussian Belief Propagation Inference with Lie Groups}
\label{sec:lie_gbp}
It is possible to use Robot Web in vector space with no modification to the GBP as introduced in Section~\ref{sec:gbp}.
However, in most realistic robotics problems, there are additional details to consider due to the state space being a robot pose with rotation and translation, where careful thought about parameterisation is needed.
The use of Lie theory is a key component of modern state estimation for robotics~\cite{Barfoot:book2017}, and is applied to algorithms such as Extended Kalman Filter, and Information Filter~\cite{Cesic:etal:Automatica2017}.

In our work, we extend the four operations of GBP to support optimisation along a differentiable manifold, with a particular focus on Lie Groups, which is both a manifold and a group.
We use concepts and the notation
defined in Sol\`{a} \etal's excellent tutorial, `A micro Lie
theory for state estimation in robotics'~\cite{Sola:etal:ARXIV2018}.

First, we define two operations on the manifold: retraction $\mathcal{R}: \mathcal{M} \times T_\mathcal{\Xbar}\mathcal{M} \rightarrow \mathcal{M}$, and its inverse $\mathcal{L}: \mathcal{M} \times \mathcal{M} \rightarrow T_\mathcal{\Xbar}\mathcal{M}$, where  $\mathcal{\Xbar}\in\mathcal{M}$ is a point on a manifold $\mathcal{M}$, and $T_\mathcal{\Xbar}\mathcal{M}$ is the tangent space of $\mathcal{M}$ at $\mathcal{\Xbar}$.
In case of a Lie Group we choose:
\begin{alignat}{3}
\mathcal{\Ybar} &= \mathcal{R}(\mathcal{\Xbar}, \vectau) 
                && \triangleq \mathcal{\Xbar} \oplus \vectau 
                && \triangleq  \mathcal{\Xbar} \circ \text{Exp}(\vectau) \in \mathcal{M}~, \\
\vectau &= \mathcal{L}(\mathcal{\Xbar}, \mathcal{\Ybar}) 
                       && \triangleq \mathcal{\Ybar} \ominus \mathcal{\Xbar} 
                       && \triangleq \text{Log}(\mathcal{\Xbar}^{-1} \circ \mathcal{\Ybar}) \in T_\mathcal{\Xbar}\mathcal{M}~,
\end{alignat}
where we use the Lie groups exponential and logarithmic map.
For brevity, we focus on Lie Groups and will use its notation for the derivation. 

Let $\mathcal{\Xbar}$ represent the point estimate of a member of a Lie Group.
A Gaussian distribution around this point estimate is represented using the tangent space as follows:
\beq
\mathcal{X} \sim \mathcal{N}(\mathcal{\Xbar}, \matLambda^{-1})
~,
\eeq
where:
\beq
\mathcal{X} = \mathcal{\Xbar} \oplus \xi \\
~,\text{and}~
\xi \sim \mathcal{N}(0, \matLambda^{-1})
~.
\eeq
The core of how we use Lie Theory within GBP is the choice that
{\em all
messages to or from a Lie Group variables take the form of a point
estimate, represented by the full over-parameterised Group element,
together with a minimal precision matrix defined in the tangent space
around that element}.
So, when variable $\mathcal{X}$ represents a transformation which  is a member of a
Lie  Group, all messages to it will take the
form $\mathcal{\Xbar}, \matLambda$, where $\matLambda$ is
a precision matrix in the tangent space at the point estimate $\mathcal{\Xbar}$.

This approach gives maximum flexibility and minimises the need for
independent devices to have knowledge or memory of each other (which
might be required with alternative ideas, such as that messages would
represent perturbations around some remembered stored element).

\subsection{Belief Update at a Variable Node}
\label{sec:belieflie}

At any stage, we can calculate a new marginal distribution at a variable node given all of the message passing that
has happened to date; this is the current estimate of the position of a robot at a particular time step.

Consider a Lie Group variable connected to
$N \geq 1$ factor nodes $i = 1 \ldots N$. In a message passing step, it must output a message to one of these
factors, e.g. $i=a$. Each of the $N-1$ incoming messages has the form
$\mathcal{\Xbar}_i, \matLambda_i$; for each message, the precision
matrix is in the local tangent space around the Lie Group element.
We transform the tangent vector with its associated Gaussian distribution into the same tangent space, such that we can combine them.
There are different possible choices for which tangent space to use, but a good choice is to use a previous estimate of the variable's state
$\mathcal{\Xbar}_0$ (calculated the last time we did a belief update at that node; see Section~\ref{sec:belieflie}) because this requires a minimal transformation of messages. To transform the mean of an incoming message into this tangent, space we perform:
\beq
  \vectau_i = \mathcal{\Xbar}_i \ominus \mathcal{\Xbar}_0
  \label{equ:transformmeantangentspace}
~.
\eeq
And to transform its precision matrix:
\beq
\matLambda_{\vectau_i} =
\Jac{\top}{r}(\vectau_i) \matLambda_i \Jac{}{r}(\vectau_i)
  \label{equ:transformprecisiontangentspace}
~,
\eeq
where $\Jac{}{r}(\vectau)$, right Jacobian of $\mathcal{X} = \text{Exp}(\vectau)$, is:
\beq
\Jac{}{r}(\vectau) =
\lim_{\epsilon\rightarrow0} \frac{\text{Exp}(\vectau + \epsilon) \ominus \text{Exp}(\vectau)}{\epsilon}
~.
\eeq

To compute the belief at a variable, first, we sum all precision matrices to determine the total precision:
\beq
\matLambda_{\vectau} = \sum^N {\matLambda_{{\vectau}_i}}
~,
\eeq
and combine the messages to obtain the total tangent vector:
\beq
\vectau =
\matLambda_{\vectau}^{-1}
\sum^N
  {\matLambda_{\vectau_i}} \vectau_i
    ~.
\eeq
Finally we apply this tangent vector to $\mathcal{\Xbar}_0$ to obtain the group element
which is the mean of the belief of the variable:
\beq
\mathcal{\Xbar} = \mathcal{\Xbar}_0 \oplus\vectau
~,
\eeq
and warp the total precision into the tangent space of the new $\mathcal{\Xbar}$:
\beq
\matLambda_{\mathcal{\Xbar}} =
\Jac{-\top}{r}(\vectau) \matLambda_{\vectau} \Jac{-1}{r}(\vectau)
~.
\eeq

\subsection{Variable to Factor Message}
Given the latest incoming messages $\mathcal{\Xbar}_i, \matLambda_i$ at the variable and the  previously calculated belief mean $\mathcal{\Xbar}_0$, we  transform all incoming messages into the tangent space of $\mathcal{\Xbar}_0$ using Equation~\eqref{equ:transformmeantangentspace} and Equation~\eqref{equ:transformprecisiontangentspace} to obtain $\vectau_i$ and $\matLambda_{\vectau_i}$.
Now that all $\vectau_i$ and $\matLambda_{\vectau_i}$ are defined in the tangent space of
$\mathcal{\Xbar}_0$,
we can add all precision  matrices to determine the total precision:
\beq
\label{equ:precisionexcludinga}
\matLambda_a = \sum_{i \neq a}^N {\matLambda_{{\vectau}_i}}
~.
\eeq
Next
we can combine the messages to obtain the tangent vector
of the outgoing message:
\beq
\label{equ:tangentvectorexcludinga}
\vectau_a =
\matLambda_a^{-1}
\sum_{i \neq a}^N
  {\matLambda_{\vectau_i}} \vectau_i
    ~.
\eeq
Finally we apply this tangent vector to $\mathcal{\Xbar}_0$ to obtain the group element
which is the mean of the outgoing message:
\beq
\mathcal{\Xbar}_a = \mathcal{\Xbar}_0 \oplus\vectau_a
~,
\eeq
and warp the total precision to the tangent space of $\mathcal{\Xbar}_a$:
\beq
\matLambda_{\mathcal{\Xbar}_a} =
\Jac{-\top}{r}(\vectau_a) \matLambda_a \Jac{-1}{r}(\vectau_a)
~.
\eeq
The outgoing message to factor $a$ is:
$\mathcal{\Xbar}_a,
\matLambda_{\mathcal{\Xbar}_a}$, together with its most recent point estimate $\mathcal{\Xbar}_0$.

\subsection{Factor Potential of a Lie Group Parameterised Function}


Here, we consider the case where at least one of the connected
variables represent a Lie Group. Now parts of $\vecx$ will be group
elements rather than simple vectors, and messages to and from those
variables take the form of group elements together with precision matrices in the tangent space of
those elements.

We first use the most recent variable states from the incoming messages
from all variables connected to the factor to form stacked state representation $\vecx_0$.
For example, for a factor connected to one $\RR^2$ and two $\SE2$
variables:
\bea
  \vecx_0 &=& \left[
    \begin{array}{cc} \mathcal{\Xbar}_0^1 \\ \mathcal{\Xbar}_0^2 \\ \mathcal{\Xbar}_0^3 \end{array} \right]
    \in \langle \RR^2, \SE2, \SE2 \rangle ~,
\eea
where $\mathcal{\Xbar}_0^{i}$ is the variable $i$'s state.
We will use $\vecx_0$ as the linearisation point for the factor, and
perform the calculations needed for message passing in the tangent
space around this point.

Our measurement function $\vech$ output and the observation may not be Euclidean vectors.
Hence, we define an error term:
\beq
\vece(\vecx) = \vecz \dminus \vech(\vecx)~,
\eeq
where we use the notation  $\dminus$ from~\cite{Sola:etal:ARXIV2018}, which is an operation on the composite manifold ($\ominus$ operation is applied to each block of composites separately). We linearise the non-linear error term $\vece$ via a first-order Taylor expansion around $\vecx_0$:
\bea
\vece(\vecx) &\approx& \vece(\vecx_0) + \matJ(\vecx \dminus \vecx_0)
\\
&=&
\label{equ:hs_taylorexpansion}
\vece(\vecx_0) + \matJ \vectau
~.
\eea
Here $\matJ$ is the Jacobian of the measurement function
with respect to the compound tangent space:
\beq
\matJ = \frac{\partial \vece}{\partial
  \vectau}|_{\vecx = \vecx_0} 
=  \rowthree{\frac{\partial \vece}{\partial
  \vectau_1}}{\frac{\partial \vece}{\partial
  \vectau_2}}{\frac{\partial \vece}{\partial
    \vectau_3}}
~.
\eeq
Identifying that $\matA = \matJ$ and $\vecb = \vece(\vecx_0)$, and substituting into Equation ~\eqref{equ:linear_factor_potential_eta},~\eqref{equ:linear_factor_potential_Lam}:
\bea
\veceta_f &=& \matJ^{\top} \matLambda_s(0 - \vece(\vecx_0))~, \\
\matLambda_f &=& \matJ^{\top} \matLambda_s \matJ~.
\eea
$\vecz_s = 0$ as it's already included in the error term to handle the group elements.

\subsection{Factor to Variable Message}
\label{sec:messagepassingfactor}

Having linearised the factor, we can now perform message passing. Consider
 our example factor connected to one $\RR^2$ and two $\SE 2$
 variables. The factors information
vector and precision matrix are partitioned as follows:
\bea
\veceta_\mu &=&
\vecthree{\veceta_{\mu_1}}{\veceta_{\mu_2}}{\veceta_{\mu_3}} \\
\matLambda' &=&
\matthree{\matLambda_{11}'}{\matLambda_{12}'}{\matLambda_{13}'}{\matLambda_{21}'}{\matLambda_{22}'}{\matLambda_{23}'}{\matLambda_{31}'}{\matLambda_{32}'}{\matLambda_{33}'}
~.
\eea

If we choose the output variable to be the third variable, we need to first
condition the factor on the incoming messages from variables 1 and 2
and then marginalise to achieve an output distribution over variable 3.
Conditioning is
achieved  by adding as follows:
\bea
\veceta_{C} &=& \vecthree{\veceta_{\mu_1} + \veceta_1}{\veceta_{\mu_2}+ \veceta_2}{\veceta_{\mu_3}} \label{eqn:etacond}\\
\matLambda_{Cs}' &=&
\matthree{\matLambda_{{11}}' +
  \matLambda_1}{\matLambda_{{12}}'}{\matLambda_{{13}}'}{\matLambda_{{21}}'}{\matLambda_{{22}}'
+
\matLambda_2}{\matLambda_{{23}}'}{\matLambda_{{31}}'}{\matLambda_{{32}}'}{\matLambda_{{33}}'} \label{eqn:lambdacond}
~,
\eea
where $(\veceta_i$, $\matLambda_i)$ describe the incoming message $(\mathcal{\Xbar}_i, \matLambda_{\mathcal{\Xbar}_i}')$ from
variable $i$ in the  tangent space of $\vecx_0$.
\bea
  \vectau_i &=& \mathcal{\Xbar}_i \dminus \vecx_0^{i} \\
  \matLambda_i' &=& \Jac{\top}{r}(\vectau_i) \matLambda_{\mathcal{\Xbar}_i}' \Jac{}{r}(\vectau_i) \\
  \eta_i &=& \matLambda_i' \vectau_i
~.
\eea
Here $\vecx_0^{i}$ represents the $i$th block in the composite.


To complete message passing, from this joint distribution we must
marginalise to obtain a distribution $\mathcal{N}^{-1}(\veceta_3, \matLambda_3)$ over the
output variable.
Eustice~\etal~\cite{Eustice:etal:ICRA2005} give the formula for marginalising a general partitioned Gaussian state in the information form.
If the joint distribution is:
\bea
    \label{equ:alpha}
    \veceta &=& \vectwo{\veceta_\alpha}{\veceta_{\beta}}\\
    \label{equ:beta}
    \matLambda &=& w
    ~,
\eea
then the marginal distribution over the variables $\alpha$, achieved by integrating over the variables $\beta$, is:
\bea
    \label{equ:emarg}
    \veceta_{M\alpha} &=& \veceta_\alpha - \matLambda_{\alpha\beta} \matLambda_{\beta\beta}^{-1} \veceta_\beta \\
    \label{equ:Lmarg}
    \matLambda_{M\alpha} &=& \matLambda_{\alpha\alpha} - \matLambda_{\alpha\beta} \matLambda_{\beta\beta}^{-1} \matLambda_{\beta\alpha} 
    ~.
\eea

The distribution $\mathcal{N}^{-1}(\veceta_3, \matLambda_3)$ is still in the tangent space of the linearised
factor, so finally we transform back to a Lie Group element with the
information matrix in its own tangent space as follows to form
the outgoing message $\mathcal{\Xbar}_{o3},
\matLambda_{\mathcal{\Xbar}_{o3}}$:
\bea
\vectau_3 &=& \matLambda_3^{-1} \veceta_3 \\
\mathcal{\Xbar}_{o3} &=& \mathcal{\Xbar}_{0}^{3} \oplus \vectau_3 \\
\matLambda_{\mathcal{\Xbar}_{o3}}
&=& \Jac{-\top}{r}(\vectau_3)
\matLambda_3
\Jac{-1}{r}(\vectau_3)
~.
\eea



\section{Robot Web: Core Design and Structure}
\label{sec:robotweb}
We   will use the term `robot' for any device involved in the Robot
Web, but some of these could be beacons, sensor nodes, or any other
type of participating entity, which could be moving or stationary.

\subsection{Partitioning of the Factor Graph}
The fundamental structure of the Robot Web is the full probabilistic
factor graph which represents the states of robots as variables and
the measurements they make, or any other information which is
available such as pose or smoothness priors, as factors. Determining estimates of the robot states
is a matter of performing inference on this factor graph to produce marginal
distributions over the variables. We will assume that all factors take
the form of Gaussian functions of the involved state variables, and
use Gaussian Belief Propagation as the mechanism for inference. Note
that GBP supports robust (heavy-tailed) factors and non-linear
measurement functions via the methods proposed in~\cite{Davison:Ortiz:ARXIV2019}, and therefore
this model is very broadly practically applicable. These are the same assumptions behind most centralised factor graph inference libraries, such as GTSAM~\cite{Dellaert:TechReport2012}, Ceres~\cite{CeresManual}, and g2o~\cite{Kummerle:etal:ICRA2011}.

The key concept of the Robot Web is to distribute responsibility for
storing and updating the full-factor graph, by dividing it up
between the robots taking part. Figure~\ref{fig:factor} illustrates this
for an elementary case of three moving robots,
each with internal odometry sensing and an outward-looking sensor able
to make observations of the other robots. We use different colours to highlight the parts of the factor graph for which each robot is responsible, and refer to the different robots using different Greek alphabets.
A Robot  $\alpha$ stores:
\begin{itemize}
\item The set of variables $\vecv_t^\alpha$ representing its state at discrete times $t$. It could store a whole history of states or a finite window. Most commonly these states will be
  multi-dimensional variables which
  directly represent robot pose, though any other aspects of internal state could be included.
We will discuss pose parametrisation in detail later in Section~\ref{sec:lie_gbp}.
\item The set of factors $f_t^\alpha$ or $f_{t, t+1}^\alpha$ representing internal, priors or
  proprioceptive measurements. Each of these factors connects to one or more of the robot's own
  state variables. Common examples would be a unary factor representing a GPS pose measurement, or a binary factor
  connecting two temporally consecutive states representing an
  odometry or inertial measurement.
\item A set of factors $g_t^{\alpha,\beta}$ representing exteroceptive
  measurements made by this robot of other robots. Specifically,
  factor $g_t^{\alpha,\beta}$ represents a measurement
  made by this robot $\alpha$ of another  robot $\beta$ at time $t$. This factor connects one state  variable $\vecv_t^\alpha$  from robot $\alpha$ with state variable $\vecv_t^\beta$ of robot $\beta$ at the corresponding time.
\end{itemize}

There is an important design choice here: factors representing
inter-robot measurements are stored by the robot making the
measurement. This is because the details of measurement factors
depend on the type and calibration of the sensor involved, and in this
way, those details only need to be known to the robot carrying the
sensor.  Note also that we assume for now that all robots have
globally synchronised clocks for timestamping of measurements (though
we will see that all computation and communication can be asynchronous).

The factor graph evolves and grows dynamically. At initialisation, a robot
will have just one variable node. As it moves, and measures its own
incremental motion with odometry or similar, it adds the appropriate
variables and factors to its internal factor graph. GBP runs continuously
on the robot's internal factor graph, producing always-updating
marginal distributions for each variable. The message passing pattern of GBP within a robot's internal factor graph is not important but should be rapid and global enough to keep the graph fragment mostly close to convergence.

\subsection{Message Passing and Communication Model}
\label{sec:comm_model}
When the robot uses an
outward-looking sensor to make an observation of the relative location
of another robot, it creates a factor for this measurement, connects it
to its current live pose variable, and the factor takes part in local GBP.
The other end of this factor
will initially be unconnected, because the appropriate variable to
attach it to is stored by another robot: the factor-to-variable edge
crosses the `dotted line' boundary (see Figure~\ref{fig:factor}), separating factor graph fragments.
When local GBP generates an outgoing message from this factor which crosses the dotted line, that message is made available to the other robot that needs it. 

The key idea behind Robot Web is that its inter-robot communication model is flexible and does not require synchronous or bidirectional communication. 
Instead, each robot can broadcast information at its own rate, which is particularly useful in large-scale systems where synchronising communication across multiple robots can be a challenge.
This is only possible as GBP can converge even with an arbitrary message schedule~\cite{Ortiz:etal:ARXIV2021} meaning
that the communication between robots can be completely asynchronous and ad-hoc, but the overall graph made up of many fragments will converge to the global estimates. Here, we detail how the robots communicate with each other in our work.

\subsubsection{Communication Model}
The asynchronous nature of the communication allows for a variety of options for message delivery, such as the publish-subscribe model used in systems like ROS/ROS2, where devices broadcast messages and listen to topics of interest, or the pull model, where devices query each other for information. In our work, we do not assume that messages will always be delivered, and any loss of messages will only result in a possible decrease in localisation accuracy, rather than causing a deadlock or critical failure. Additionally, we stress that our approach does not involve any shared global information among the robots. Each robot only exchanges messages with the others and does not share sensor models, initialisation status, or even the number of robots participating in the optimisation process.
This combination of asynchronous communication and lack of shared global information allows the system to function even if there are fewer communication rounds than the total number of robots.


\subsubsection{Inter-device Factor Discovery}
In Robot Web, the connection over the inter-device factors is formed lazily.
A factor is created when an observation occurs. However, it takes a few iterations of message passing before the factor can be linearised. Specifically, when Robot $\alpha$ observes Robot $\beta$ at timestamp $t$, a factor $g_t^{\alpha, \beta}$ is formed. As the observation is made by Robot $\alpha$, it owns the factor.
Robot $\alpha$ publishes an empty message $\vecm_{g_t^{\alpha, \beta} \rightarrow \vecv_t^\beta}$.
Upon Robot $\beta$ receiving the message, it will publish the message $\vecm_{ \vecv_t^\beta \rightarrow g_t^{\alpha, \beta}}$ together with the linearisation point $\mathcal{\Xbar}_t^\beta$.
When Robot $\alpha$ receives this message, it linearises the factor and starts the optimisation process. In the succeeding rounds, Robot $\beta$ will receive a message from Robot $\alpha$ which it can use to refine its pose estimate.

As GBP performs operations locally and does not use global information (e.g. topology of the graph), this sequence of message exchange occurs asynchronously, and GBP continues optimising as it discovers new inter-device factor connections.

\subsubsection{Robot Web Page Interface}
One of the main motivations for the design choices made in Robot Web is the desire for {\bf distributed scalability}. By providing a uniform interface, the robots can be added to or removed from the Robot Web in a fully dynamic manner, or can freely change their internal methods or software as long as they maintain the same interface.
 The internal complexity of each robot's processing may be slightly
increased because of this, but this is a small price to pay for global scalability.
 
To achieve this goal, one potential approach is to use a simple Web protocol (e.g. HTTP) for all inter-robot communication, with each robot hosting outgoing messages as a Web page. This allows inter-robot communication to happen in arbitrary patterns and in a read-only style, which can contribute to the scalability and flexibility of the system.

\section{Demonstrations and Experiments in a Simulated Environment}
\label{sec:demos}

We present extensive simulation demonstrations of Robot Web localisation for the case of many robots
with planar 2D motion and noisy odometry and inter-robot range-bearing
measurements. Our simulation uses metric units and models an application like a warehouse setting where tens or hundreds of robots roam through an environment 100m across with randomly generated paths. 
Usually, we add a handful of known beacon landmarks to the environment, whose positions are known in advance to all robots, but are widely spread so that robot-landmark measurements are much less frequent than robot-robot measurements. 
The main role of the landmarks is to anchor the whole web to an absolute coordinate frame over long periods of operation.

Our simulation uses a fully distributed program structure equivalent to what could
be achieved on a true multi-robot system.

\subsection{Implementation Details}
In our experiments, we run the robots in a square arena of width 100m with 10 known beacons where all robots move through 100 pose steps. 
All variable nodes in the current simulation are represented using $\SE2$, and three different factors are implemented:

\textbf{Anchor Factor:}
If needed, we can use unary anchor factors which are priors on the poses of robots before they start moving.
In most experiments, we use these factors to represent fairly well-known initial robot positions at the start of motion, though note that in Section~\ref{sec:dynamic} we show that new robots can be added to an existing web without any pose priors.

An anchor factor is defined as $\vech(\vecx) = \vecx$ with measurement $\vecz \in \SE2$,
which is the prior pose estimate. The uncertainty assigned to the anchor factors  in our main experiments  is:
$\sigma_x = 0.1$m, $\sigma_y = 0.1$m, and $\sigma_{\theta} = 0.01$ rad.

\textbf{Odometry Factor:}
For robot odometry, a binary factor $\vech(\vecx_1, \vecx_2) = \vecx_1 \ominus \vecx_2$ with measurement $\vecz \in \se2$
represents a relative pose measurement. The uncertainty assigned to odometry factors per metre step is:
$\sigma_x = 0.1$m, $\sigma_y = 0.01$m, and $\sigma_{\theta} = 0.01$ rad. As each robot moves along its x-axis, the uncertainty on $x$ is higher than $y$.

\textbf{Range-Bearing Factor:}
We use a range-bearing sensor for the measurements between robots, or between robots and landmarks. The measurement function is $\vech(\vecx_1, \vecx_2) = (r, b)$, where $r$ is the
Euclidean distance between $\vecx_1, \vecx_2$ and $b$ is the angle between the $\vecx_1, \vecx_2$ in the coordinate frame of $\vecx_1$.
The range-bearing measurement is defined as $z \in \langle \RR, \SO2\rangle$.
By default the uncertainty assigned to range/bearing factors is: $\sigma_r = 0.01$m, $\sigma_b = 0.05$ rad, with the sensor range limited to 30m. DSC~\cite{Agarwal:etal:ICRA2013} is used as the robust kernel with $\Phi = 10$.
Alongside the Gaussian noise, to 10\% of all range-bearing measurements, we additionally add a huge amount of uniform noise: $r_n\sim\mathcal{U}(0, 30)$, $b_n\sim\mathcal{U}(0, \pi)$ to simulate non-Gaussian noise which makes these measurements essentially useless. 

We use a communication pattern which simulates a limited peer-to-peer
communication budget, where each robot connects to and reads the Robot Web page from other robots in a sequential, random pattern with closer
robots are more likely to be selected.  The idea is that this is similar
to a robot sequentially switching its Wi-Fi connection between peers
with strong signals.

We generated a  noisy distance sample between Robot $\alpha$ and Robot $\beta$ as $d_{\alpha, \beta}\sim\mathcal{N}(\dbar_{\alpha, \beta}, 0.1)$, where
$d_{\alpha, \beta}$ is the random sample and $\dbar_{\alpha, \beta}$ is the ground truth distance between Robot $\alpha$ and Robot $\beta$.
We define the neighbourhood of $\alpha$, $N(\alpha)$, as the set of robots which Robot $\alpha$ can communicate with.
The probability $C_{\alpha, \beta}$ that Robot $\alpha$ communicates with Robot $\beta \in N(\alpha)$ is:
\beq
\label{eqn:communication}
p(C_{\alpha, \beta}) = \frac{1 / d_{\alpha, \beta}^2}{\sum_{\omega \in N(\alpha)}{1/d_{\alpha, \omega}^2}}
~.
\eeq
In this work, we assume that $N(\cdot)$ includes all robots. 
Each robot performs 20 iterations of GBP per movement step and at each GBP step robots communicate with only one neighbour.
Factors are dampened~\cite{Murphy:etal:1999} by 0.2, and the factors linearise at every iteration.
Any changes to the default parameters will be specified in the individual experiments.

\subsection{Convergence and Computational Properties}

\begin{figure*}[!t]
  \begin{subfigure}[b]{0.32\textwidth}
    \includegraphics[width=\textwidth]{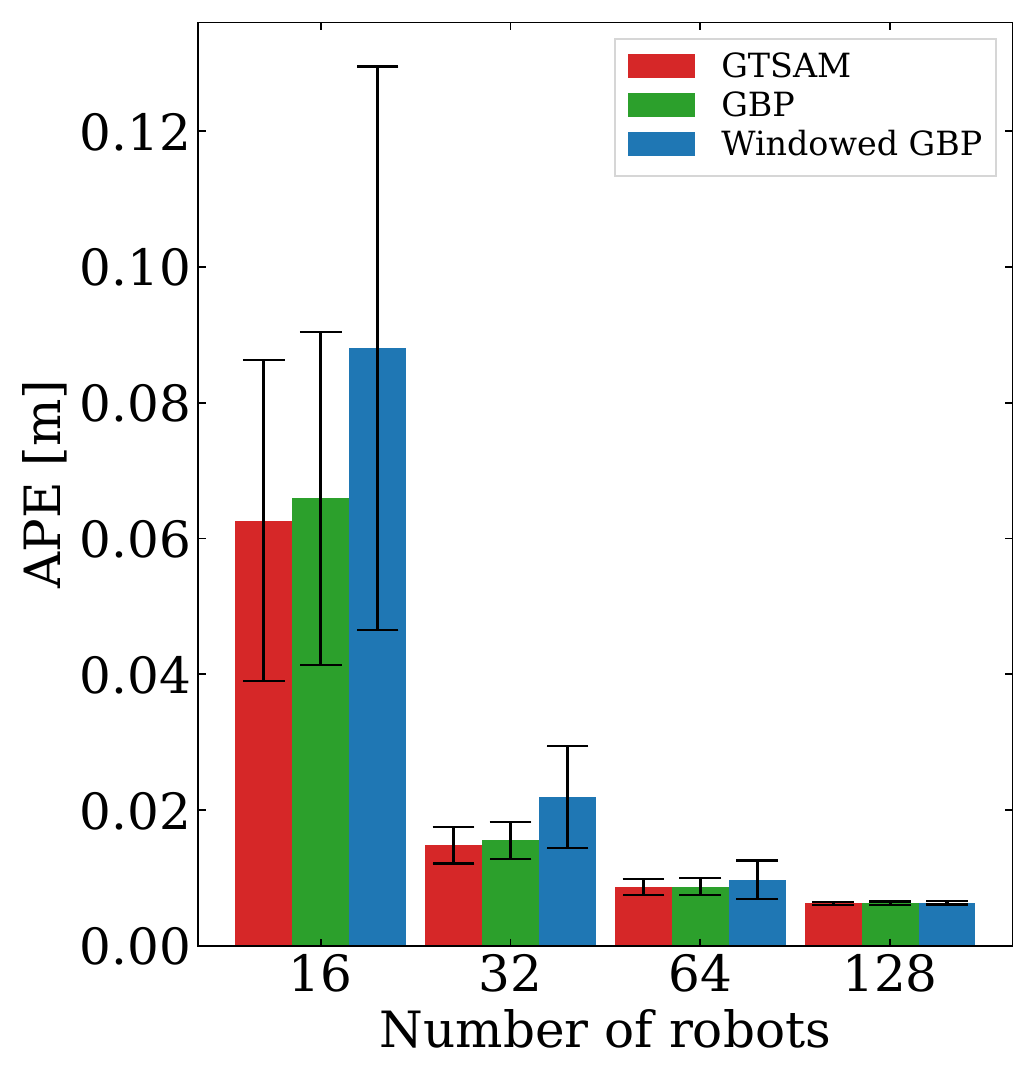}
    \caption{Mean RMSE ATE}\label{fig:batchsolvercomp_ATE}
  \end{subfigure}
  \hfill
  \begin{subfigure}[b]{0.32\textwidth}
    \includegraphics[width=\textwidth]{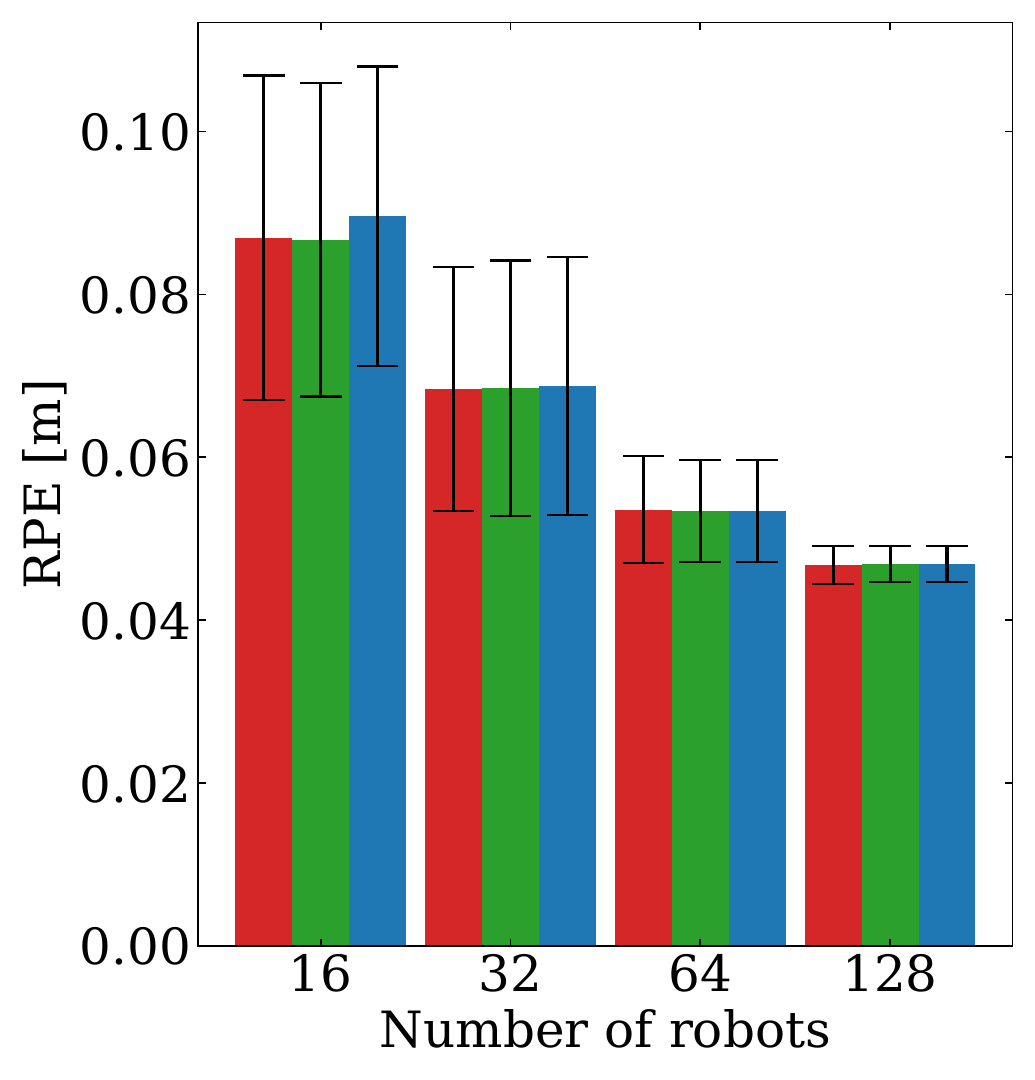}
    \caption{Mean RMSE Translational RPE}\label{fig:batchsolvercomp_RPE_t}
  \end{subfigure}
  \hfill
  \begin{subfigure}[b]{0.32\textwidth}
    \includegraphics[width=\textwidth]{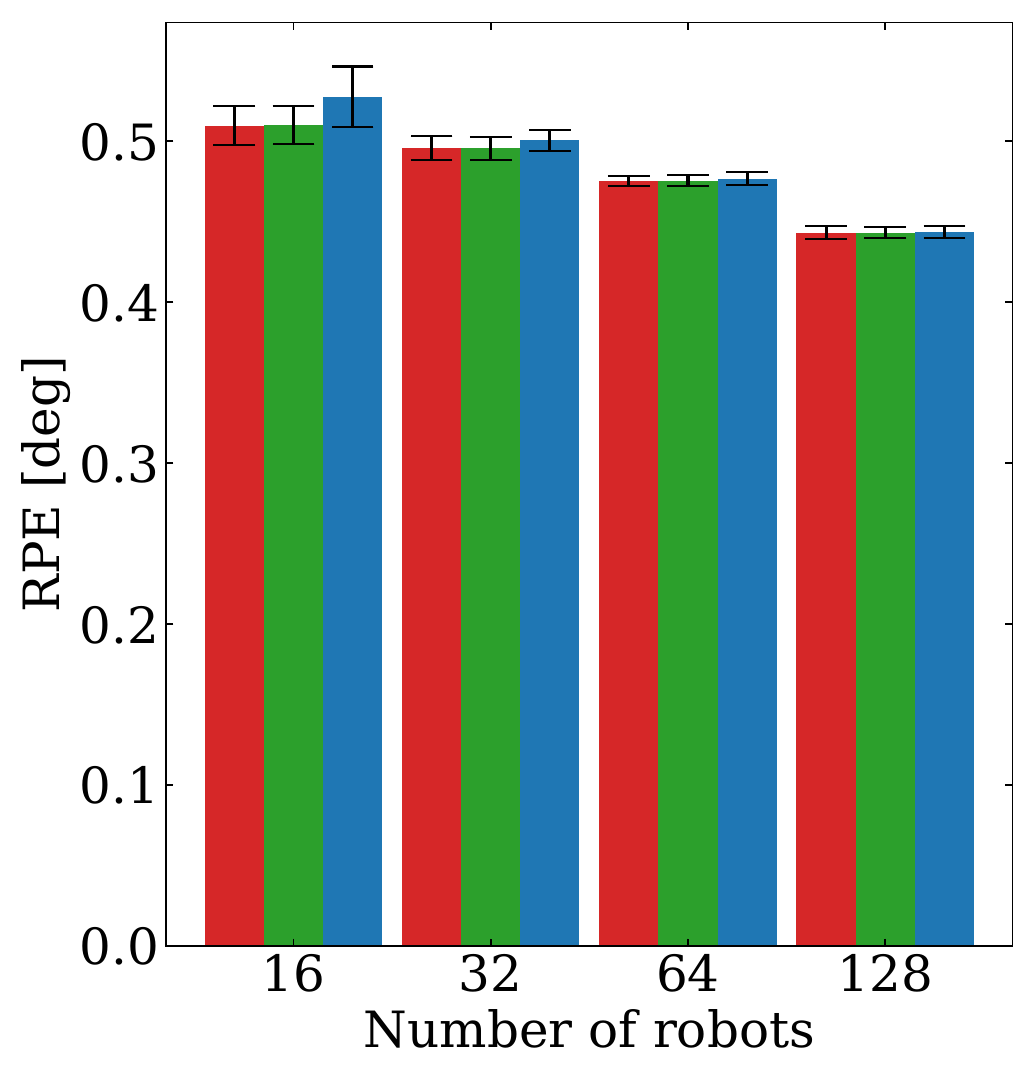}
    \caption{Mean RMSE Rotational RPE}\label{fig:batchsolvercomp_RPE_R}
  \end{subfigure}
  \hfill
  \caption{
    In a simulated environment, N robots are moving around in an environment with 10 known landmarks for 100 poses each.
    GTSAM optimises the factor graph after every pose insertion rather than solving after all poses are inserted to keep
    the comparison fair. GBP uses the full factor graph to optimise, while Windowed GBP only uses only the last 5 poses.
    The results are the average of 10 runs with different random initialisation, and the error bar represents one standard deviation of uncertainty.
  }
\label{fig:batchsolvercomp}
\vspace{2mm} \hrule
\end{figure*}
\begin{figure}[!t]
    \includegraphics[width=\columnwidth]{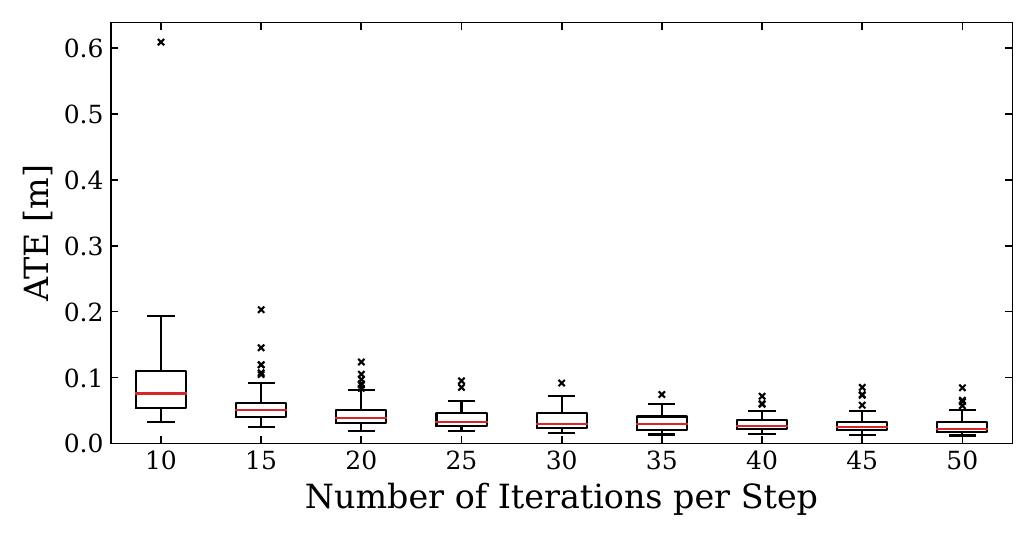}
  \caption{\label{fig:vary_iter}
  Increasing the number of iterations per step decreases the overall error. Even with a small number of iterations, GBP is able to provide good localisation, which can be further refined by increasing the iterations. 
  The red line shows the median, the box extends from the first quartile to the third quartile, the whisker extends from the box by 1.5 inter-quartile range, and the outliers are marked with a cross.
  In a simulated environment, 50 robots are moving in an environment with 10 known landmarks for 100 poses each.
  Each result is a summary of 50 runs with different random initialisation.
  }
\vspace{2mm} \hrule
\end{figure}
\begin{table}[t]
\caption{\label{table:ate}
The RMSE ATE of the trajectories for different numbers of robots in simulation. We report the mean error and the standard deviation of 10 runs with different random initialisation.
}

\centering
\setlength\tabcolsep{4pt}
\begin{tabular}{cccccc}
\toprule

N & Range & Noise & GTSAM & GBP & Windowed \\
  &    [m]       & [m, rad] & $\mu\pm\sigma$ [m] & $\mu\pm\sigma$ [m] & $\mu\pm\sigma$ [m] \\
\midrule
\midrule
16 & 10 & 0.01, 0.05 & $0.660 \pm 0.217$ & $0.770 \pm 0.183$ & $0.934 \pm 0.152$ \\
 & 10 & 0.05, 0.1 & $0.690 \pm 0.218$ & $0.773 \pm 0.192$ & $0.975 \pm 0.127$ \\
 & 30 & 0.01, 0.05 & $0.063 \pm 0.024$ & $0.066 \pm 0.025$ & $0.088 \pm 0.042$ \\
 & 30 & 0.05, 0.1 & $0.081 \pm 0.019$ & $0.087 \pm 0.021$ & $0.117 \pm 0.040$ \\
 \midrule
32 & 10 & 0.01, 0.05 & $0.314 \pm 0.062$ & $0.462 \pm 0.080$ & $0.561 \pm 0.076$ \\
 & 10 & 0.05, 0.1 & $0.344 \pm 0.049$ & $0.437 \pm 0.058$ & $0.597 \pm 0.063$ \\
 & 30 & 0.01, 0.05 & $0.015 \pm 0.003$ & $0.016 \pm 0.003$ & $0.022 \pm 0.008$ \\
 & 30 & 0.05, 0.1 & $0.035 \pm 0.003$ & $0.036 \pm 0.004$ & $0.043 \pm 0.006$ \\
 \midrule
64 & 10 & 0.01, 0.05 & $0.154 \pm 0.018$ & $0.290 \pm 0.072$ & $0.358 \pm 0.072$ \\
 & 10 & 0.05, 0.1 & $0.181 \pm 0.023$ & $0.256 \pm 0.064$ & $0.375 \pm 0.080$ \\
 & 30 & 0.01, 0.05 & $0.009 \pm 0.001$ & $0.009 \pm 0.001$ & $0.010 \pm 0.003$ \\
 & 30 & 0.05, 0.1 & $0.023 \pm 0.001$ & $0.023 \pm 0.001$ & $0.024 \pm 0.003$ \\
 \midrule
128 & 10 & 0.01, 0.05 & $0.060 \pm 0.004$ & $0.105 \pm 0.016$ & $0.134 \pm 0.015$ \\
 & 10 & 0.05, 0.1 & $0.082 \pm 0.004$ & $0.102 \pm 0.009$ & $0.158 \pm 0.011$ \\
 & 30 & 0.01, 0.05 & $0.006 \pm 0.000$ & $0.006 \pm 0.000$ & $0.006 \pm 0.000$ \\
 & 30 & 0.05, 0.1 & $0.016 \pm 0.000$ & $0.016 \pm 0.000$ & $0.016 \pm 0.000$ \\
\bottomrule%
\end{tabular}
\vspace{2mm}
\end{table}

\begin{table}[t]
\caption{\label{table:translation_rpe}
The translational RMSE RPE of the trajectories for different numbers of robots in simulation. We report the mean error and the standard deviation of 10 runs with different random initialisation.
}
\centering
\setlength\tabcolsep{3pt}
\begin{tabular}{cccccc}
\toprule

N & Range & Noise & GTSAM & GBP & Windowed \\
  &    [m]       & [m, rad] & $\mu\pm\sigma$ [m] & $\mu\pm\sigma$ [m] & $\mu\pm\sigma$ [m] \\
\midrule
\midrule
16 & 10 & 0.01, 0.05 & $0.321 \pm 0.084$ & $0.349 \pm 0.071$ & $0.382 \pm 0.064$ \\
 & 10 & 0.05, 0.1 & $0.336 \pm 0.082$ & $0.356 \pm 0.077$ & $0.396 \pm 0.064$ \\
 & 30 & 0.01, 0.05 & $0.087 \pm 0.020$ & $0.087 \pm 0.019$ & $0.090 \pm 0.018$ \\
 & 30 & 0.05, 0.1 & $0.119 \pm 0.025$ & $0.118 \pm 0.025$ & $0.120 \pm 0.025$ \\
 \midrule
32 & 10 & 0.01, 0.05 & $0.207 \pm 0.028$ & $0.235 \pm 0.041$ & $0.264 \pm 0.048$ \\
 & 10 & 0.05, 0.1 & $0.227 \pm 0.032$ & $0.238 \pm 0.034$ & $0.276 \pm 0.039$ \\
 & 30 & 0.01, 0.05 & $0.068 \pm 0.015$ & $0.068 \pm 0.016$ & $0.069 \pm 0.016$ \\
 & 30 & 0.05, 0.1 & $0.102 \pm 0.021$ & $0.101 \pm 0.021$ & $0.102 \pm 0.020$ \\
 \midrule
64 & 10 & 0.01, 0.05 & $0.164 \pm 0.015$ & $0.198 \pm 0.031$ & $0.216 \pm 0.033$ \\
 & 10 & 0.05, 0.1 & $0.191 \pm 0.015$ & $0.197 \pm 0.020$ & $0.227 \pm 0.030$ \\
 & 30 & 0.01, 0.05 & $0.054 \pm 0.007$ & $0.053 \pm 0.006$ & $0.053 \pm 0.006$ \\
 & 30 & 0.05, 0.1 & $0.078 \pm 0.009$ & $0.077 \pm 0.009$ & $0.077 \pm 0.009$ \\
 \midrule
128 & 10 & 0.01, 0.05 & $0.092 \pm 0.009$ & $0.105 \pm 0.009$ & $0.109 \pm 0.009$ \\
 & 10 & 0.05, 0.1 & $0.120 \pm 0.004$ & $0.126 \pm 0.003$ & $0.135 \pm 0.005$ \\
 & 30 & 0.01, 0.05 & $0.047 \pm 0.002$ & $0.047 \pm 0.002$ & $0.047 \pm 0.002$ \\
 & 30 & 0.05, 0.1 & $0.068 \pm 0.003$ & $0.068 \pm 0.003$ & $0.068 \pm 0.003$ \\
\bottomrule%
\end{tabular}
\vspace{2mm}
\end{table}

\begin{table}[t]
\caption{\label{table:rotational_RPE}
The rotational RMSE RPE of the trajectories for different numbers of robots in simulation. We report the mean error and the standard deviation of 10 runs with different random initialisation.
}
\centering
\setlength\tabcolsep{3pt}
\begin{tabular}{cccccc}
\toprule

N & Range & Noise & GTSAM & GBP & Windowed \\
  &    [m]       & [m, rad] & $\mu\pm\sigma$ [deg] & $\mu\pm\sigma$ [deg] & $\mu\pm\sigma$ [deg] \\
\midrule
\midrule
16 & 10 & 0.01, 0.05 & $0.626 \pm 0.028$ & $0.642 \pm 0.025$ & $0.776 \pm 0.036$ \\
 & 10 & 0.05, 0.1 & $0.639 \pm 0.031$ & $0.650 \pm 0.030$ & $0.789 \pm 0.033$ \\
 & 30 & 0.01, 0.05 & $0.510 \pm 0.012$ & $0.510 \pm 0.012$ & $0.528 \pm 0.019$ \\
 & 30 & 0.05, 0.1 & $0.535 \pm 0.012$ & $0.535 \pm 0.013$ & $0.579 \pm 0.015$ \\
\midrule

32 & 10 & 0.01, 0.05 & $0.574 \pm 0.011$ & $0.592 \pm 0.015$ & $0.732 \pm 0.034$ \\
 & 10 & 0.05, 0.1 & $0.589 \pm 0.011$ & $0.596 \pm 0.012$ & $0.750 \pm 0.019$ \\
 & 30 & 0.01, 0.05 & $0.496 \pm 0.007$ & $0.496 \pm 0.007$ & $0.501 \pm 0.007$ \\
 & 30 & 0.05, 0.1 & $0.519 \pm 0.007$ & $0.519 \pm 0.007$ & $0.545 \pm 0.009$ \\
\midrule

64 & 10 & 0.01, 0.05 & $0.549 \pm 0.007$ & $0.567 \pm 0.013$ & $0.695 \pm 0.034$ \\
 & 10 & 0.05, 0.1 & $0.569 \pm 0.009$ & $0.573 \pm 0.008$ & $0.711 \pm 0.029$ \\
 & 30 & 0.01, 0.05 & $0.475 \pm 0.003$ & $0.476 \pm 0.003$ & $0.477 \pm 0.004$ \\
 & 30 & 0.05, 0.1 & $0.506 \pm 0.003$ & $0.506 \pm 0.003$ & $0.517 \pm 0.003$ \\
\midrule

128 & 10 & 0.01, 0.05 & $0.518 \pm 0.004$ & $0.523 \pm 0.004$ & $0.576 \pm 0.006$ \\
 & 10 & 0.05, 0.1 & $0.540 \pm 0.004$ & $0.542 \pm 0.004$ & $0.614 \pm 0.006$ \\
 & 30 & 0.01, 0.05 & $0.443 \pm 0.004$ & $0.443 \pm 0.004$ & $0.444 \pm 0.004$ \\
 & 30 & 0.05, 0.1 & $0.490 \pm 0.002$ & $0.490 \pm 0.002$ & $0.495 \pm 0.002$ \\
\bottomrule%

\end{tabular}
\vspace{2mm}
\end{table}

A key property of our method is that the marginal estimates generated by message passing with a fixed computation and communication budget on our ever-changing factor graph may not necessarily be at complete convergence during live operation, though that is often not a problem if useful robot pose estimates are still achieved.
Nevertheless, here we show that when enough computation and communication are regularly applied, the localisation results are convergent and estimates as accurate as a batch solution on a centralised processor can be achieved. Importantly, this can be achieved  with highly efficient, realistic settings for distributed GBP computation and communication.

Here we present an experiment
to compare the accuracy of distributed Robot Web GBP against a
centralised solution of the same factor graph using GTSAM~\cite{Dellaert:AR2021}.
GBP linearises at every 5 iterations and is allowed to optimise for 20 iterations per step. To keep the comparison simple, robust kernels are not applied for both GBP and GTSAM, and we do not add uniform noise to the range-bearing measurements.

We present results for general GBP, where each robot keeps a full history of pose variables, and
Windowed GBP, where each robot maintains a sliding window of its most recent 5 poses and only processes messages relating to these.
Using a sliding window allows the average size of the factor graph, and the amount of computation needed, to remain fixed.
This allows the system to operate over an arbitrarily long period while maintaining constant computational cost.
What we lose by doing this is the possibility to improve estimates of older variables in the graph using new observations.

In these experiments, for both versions of GBP,
all robots are allowed to communicate with each other on every iteration.

We report the Root Mean Square Absolute Trajectory Error (RMSE ATE), and Root Mean Square Relative Pose Error (RMSE RPE)~\cite{Sturm:etal:IROS2012}, averaged over 10 runs with randomised robot motions in Figure~\ref{fig:batchsolvercomp}, for varying numbers of robots in the area. 
Both ATE and RPE are computed over the full trajectory.
We see that GBP and GTSAM have similar ATE across all evaluations, with only a small loss  of accuracy for GBP when the  number of robots is low. 

As the number of robots increases, the difference in ATE across the different approaches becomes negligible.
Windowed GBP reaches comparable accuracy to GTSAM and  GBP, even with a significantly smaller
computational cost when compared to full GBP optimisation.

A similar pattern is observed when we vary the sensor noise and range. As we increase the number of robots, the difference in error across different approaches reduces.
The sensor noise is increased to $\sigma_r = 0.05$m, $\sigma_b= 0.1$ rad, and the range is limited to 10m. Table~\ref{table:ate},~\ref{table:translation_rpe},~\ref{table:rotational_RPE} summarises the result of sweeps over the different parameters.

When the sensor range is limited or the number of robots is small, fewer observations are made and robots will drift more from their correct trajectories. 
Being a local algorithm, GBP can rapidly optimise local, high-frequency component errors in the network~\cite{Davison:Ortiz:ARXIV2019}, while it requires more iterations for information to propagate across the graph to optimise lower frequency component errors, such as longer drifts. 
This property is observable in Figure~\ref{fig:batchsolvercomp} where the difference in ATE between GTSAM and GBP for 16 robots is noticeable due to the low-frequency noise. Since the range of the sensors is limited, fewer observations are made when the total number of robots is small, as the arena has a lower density of robots.
Similarly, as shown in Table~\ref{table:ate} for N=128. When the sensor range is 30m, GTSAM and GBP achieve the same ATE. However, when the sensor range is 10m, a small difference exists.
However; in both cases for the relative metric RPE, similar performance is achieved even with a small number of robots, demonstrating that the local/high-frequency component errors are correctly smoothened. 

Increasing the number of iterations improves convergence as more messages are exchanged. We can verify this in Figure~\ref{fig:vary_iter}, where we vary the number of iterations per step between 10-50. We use 50 robots and report the average over 50 different runs. We see that as the number of iterations per step increases, ATE decreases; however, with diminishing returns. The optimal number of iterations per step depends on many factors (e.g. topology of the graph, communication pattern) and is an interesting direction for further research.

\subsection{Operation with a Large Number of Agents}
In terms of computational performance, it would not be meaningful to
report the speed of our C++ CPU simulation of the Robot Web algorithm,
which is designed to be fully distributed across a large number of
devices. However, in fact, our simulation 
can run in real-time on  a laptop for problems involving 100 robots or beyond using Windowed GBP, in particular, because
it is designed to take advantage of CPU parallelism using OpenMP.

\begin{figure}[!t]
    \includegraphics[width=\columnwidth]{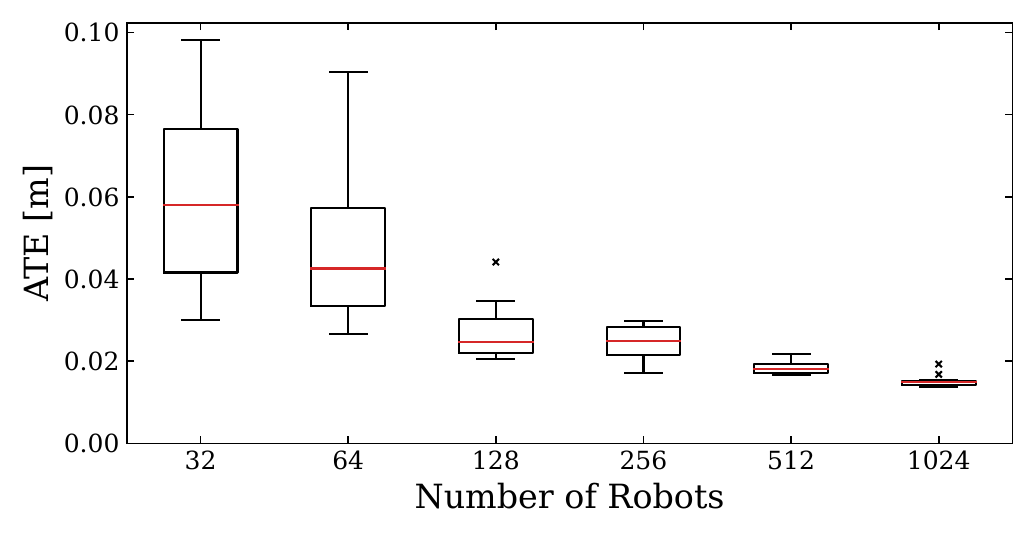}
  \caption{\label{fig:scalability}
Extreme scaling: in a simulated environment, we increase the number of robots in the  arena  to over  1000, with each robot communicating with only one other per iteration of Windowed GBP, and therefore having a  per-robot bounded computation and communication workload. 
The average ATE in  all robots' poses continues to decrease  as we increase  the number of robots.
Each result is a summary of 10 runs with different random initialisation.
  }
\vspace{2mm} \hrule
\end{figure}

Instead, we present an experiment which demonstrates the scaling properties of Windowed  GBP in a mode where the computation and communication work {\em per robot} is bounded. Figure~\ref{fig:scalability} shows the average ATE of all robot pose estimates as the number of interacting robots in our arena is raised from 32 to 1024. The result is an average of over 10 different runs. Each robot measures nearby robots but is allowed to communicate with {\em one other robot} sampled based on Equation~\eqref{eqn:communication} per GBP iteration. Robot Web handles this extreme packing and scaling straightforwardly, and the ATE for all robots continues to decrease as robots are added due to the favourable high inter-connectedness of the whole graph, despite the minimal communication allowed. These results indicate the true potential of Robot Web methods towards very high numbers of simple interacting devices.

\subsection{Operation with Outlier Measurements and Robust Factors}
\label{sec:outlier}
\begin{figure}[!tbp]
  \center
  \includegraphics[width=\linewidth]{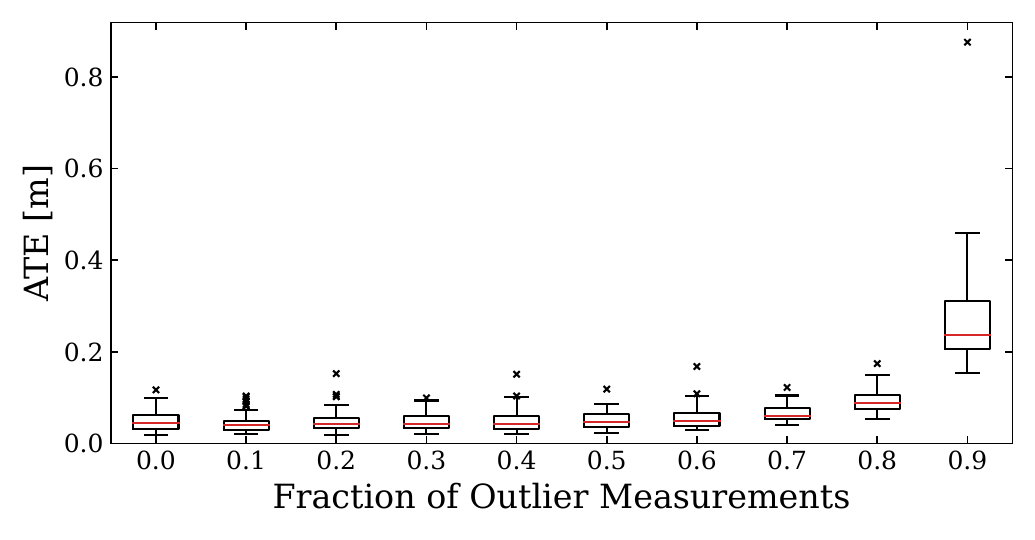}
  \caption{
        Robust factors enable remarkable resilience to a large fraction of
        outlier inter-robot sensor measurements,  with ATE remaining low up to 70--80\% of corrupt measurements to which a large amount of uniform noise is added.
      In a simulated environment, 50 robots are moving in an environment with 10 known landmarks for 100 poses each.
      Each result is a summary of 50 runs with different random initialisation.
    }\label{fig:outliermeasurements}
\vspace{2mm} \hrule
\end{figure}

Here, we demonstrate the robustness of GBP using the method for handling robust factors from FutureMapping 2~\cite{Davison:Ortiz:ARXIV2019} and the robust
kernel from
DCS~\cite{Agarwal:etal:ICRA2013}. 50 robots are used, each with a sliding window of 5.
In Figure~\ref{fig:outliermeasurements}, we show what happens to the ATE when we increase the fraction of the range-bearing measurements containing the uniform noise. We see that a huge fraction of up to 80\% of measurements can be completely corrupted but still handled by the robust measurement kernel  with very little effect on the overall accuracy of the network. This again shows the advantage of the heavily inter-connected network which GBP allows us to efficiently and incrementally optimise in a distributed manner. In this network, each pose estimate is highly over-constrained, and this is what allows the robust kernel to weed out outlier measurements.

\subsection{Operation with Unreliable Communication}
\label{sec:unreliablecomm}
\begin{figure}[!tbp]
  \center
  \includegraphics[width=\linewidth]{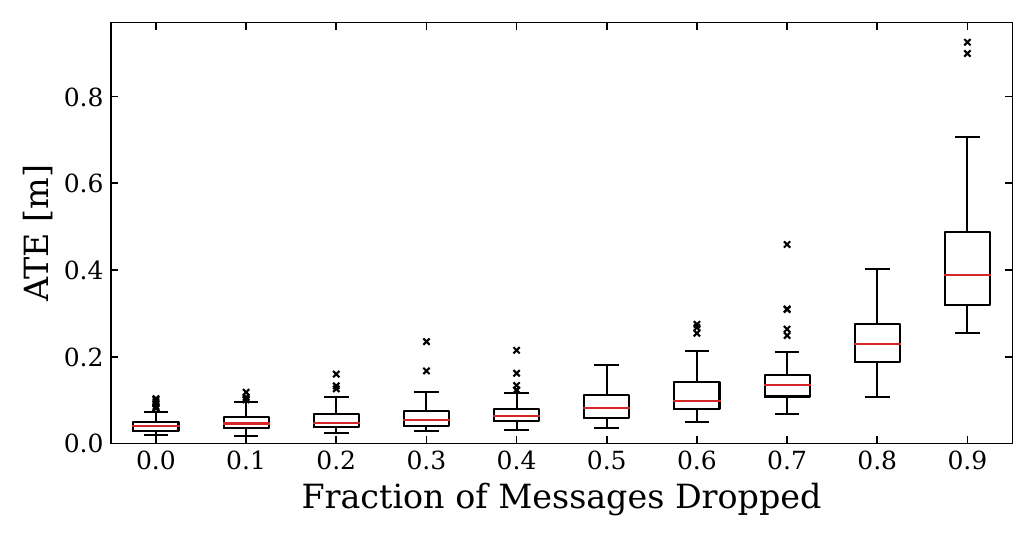}
    \caption{
      Robot Web is highly robust to a high fraction of randomly dropped messages.
      In a simulated environment, 50 robots are moving in an environment with 10 known landmarks for 100 poses each.
      The result is a summary of 50 runs with different initialisation.
    }\label{fig:dropout}
\vspace{2mm} \hrule
\end{figure}
\begin{figure}[!tbp]
  \center
  \includegraphics[width=\linewidth]{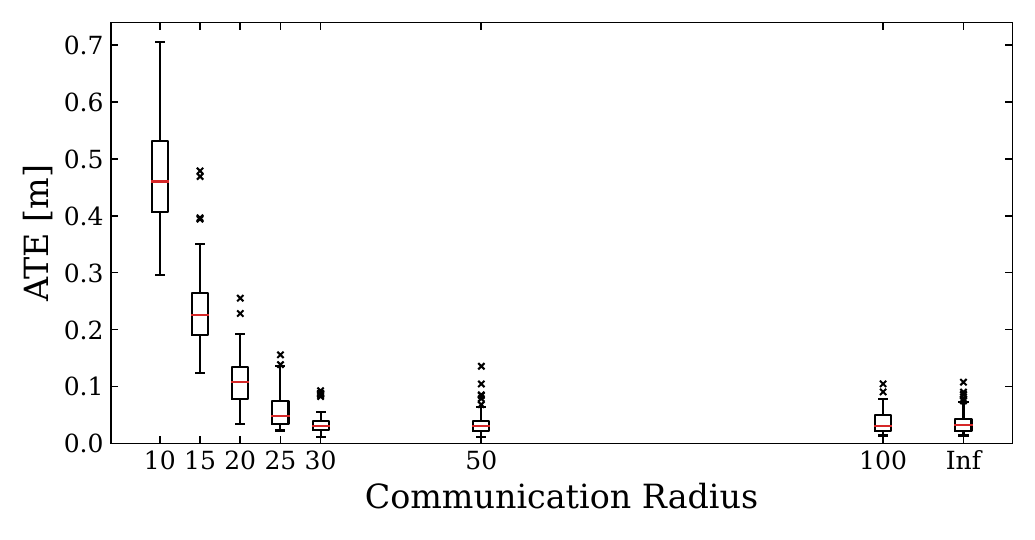}
    \caption{
      Analysis of the effect of varying the allowed communication range. `Inf' means all robots are allowed to communicate with any other robot. While increasing the communication radius improves the performance, 30m onwards, the difference is negligible.
      In a simulated environment, 50 robots are moving in an environment with 10 known landmarks for 100 poses each.
      The result is a summary of 50 runs with different initialisation.
    }\label{fig:comm_range}
\vspace{2mm} \hrule
\end{figure}

In multi-robot systems, another potential problem is the reliability of the communications. Often robots will communicate with best-effort, meaning messages can get lost in the network.
Robustness against such data loss can be challenging; however, GBP is not significantly affected, as the message scheduling can be random.
Here, we imagine that data transmission is quantised at the level of individual messages, as it might be with certain types of communication technology, and experiment to see the effect of the loss of a random fraction of messages between robots.

In Figure~\ref{fig:dropout}, we force the network to drop the messages randomly with a fixed probability which we gradually increase and investigate how that affects ATE.
For example, if Robot~$\alpha$ sends 3 rows of message $\{M_1, M_2, M_3\}$ to Robot~$\beta$, the network may drop $M_2$, and Robot $\beta$ will only receive  $\{M_1, M_3\}$.
In this experiment, we also see very advantageous properties for GBP, which retains a low ATE up to at least 50\% message loss in this setting.

We further evaluate the effect of poor communication in terms of communication range. The communication radius of the robots was adjusted to range from 10m to 100m and an infinite radius. In line with previous experiments, each robot communicates with only one other robot per iteration. We disable the sliding window and perform a full pose update for this evaluation such that robots can exchange messages asynchronously on rendezvous.

As shown in Figure~\ref{fig:comm_range}, reducing the communication range decreases the performance; however, beyond a radius of 25m, the performance improvements are minimal.
In this configuration, there may be robots who never communicate with one another though they've made measurements of each other. As GBP has no synchronisation, such cases are simply ignored without the need for specific procedures.

\subsection{Operation Under Poor Initialisation}
\begin{figure}[!tbp]
  \center
  \includegraphics[width=\linewidth]{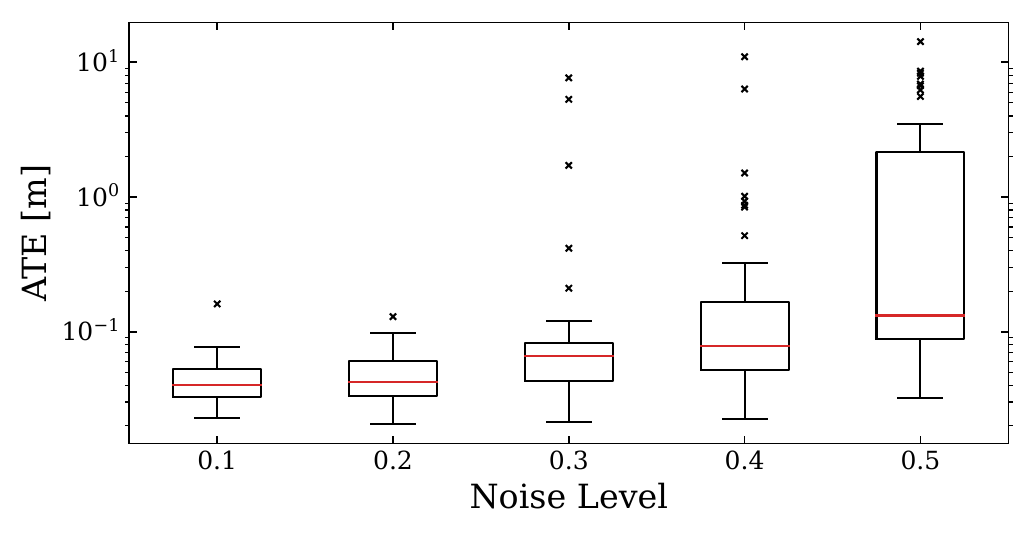}
    \caption{
      Robot Web demonstrates its resilience to large initialisation errors. We add to the initial pose a noise sampled from a Gaussian with a standard deviation of ($n$ m, $n$ m, $n$ rad), where $n$ represents the noise level. Note that the graph is plotted on a logarithmic scale.
      In a simulated environment, 50 robots are moving in an environment with 10 known landmarks for 100 poses each.
      The result is a summary of 50 runs with different initialisation.
    }\label{fig:prior_noise}
\vspace{2mm} \hrule
\end{figure}

The initialisation is important for multi-robot localisation, especially for handling outlying measurements. 
However, good initialisation may not always be available in the real world. Here, we analyse the effect of increasing the noise on the initialisation and when the system breaks.
We vary the $(\sigma_x, \sigma_y, \sigma_\theta)$ from (0.1m, 0.1m, 0.1 rad) to (0.5m, 0.5m, 0.5 rad). The percentage of outlier range-bearing sensor measurements remains to be fixed at 10\%.

We plot the graph on a logarithmic scale for clarity but notice that at noise level (0.5m, 0.5m, 0.5 rad), the value of the upper whisker is 3.51m whereas at (0.4m, 0.4m, 0.4 rad) it is 0.32m, clearly showing that the error explodes.
Initialisation is critical for outlier rejection, and with a poor initialisation, good observations will have high energy and possibly lie in the outlier region of the robust kernel, making the optimisation problem challenging.
While our approach demonstrates robustness against up to a large initialisation noise of (0.2m, 0.2m, 0.2 rad), improving the robustness to poor initialisation is an interesting direction for future works.

\begin{table*}[t]
\caption{
\label{table:comparison_dsolvers}
A comparison of the different distributed solvers for solving multi-robot pose-graph optimisation. We report the initial cost, the solution of centralised Gauss-Newton (GN), and the cost and the number of iterations required for convergence for the different distributed solvers: distributed Block Gauss-Seidel  (DGS) and distributed Block Jacobi Method from~\cite{Choudhary:etal:IJRR2017} and ours. Across all datasets, though distributed, our method and DGS obtains similar cost to the centralised GN.
}
\centering
\begin{tabular}{ccc|cccccc}
\toprule
Dataset & Initial Cost & Centralised GN & \multicolumn{2}{c}{Block Gauss-Seidel} & \multicolumn{2}{c}{Block Jacobi Method} & \multicolumn{2}{c}{Ours} \\
        &         &             & \#Iter & Cost            & \#Iter & Cost  & \#Iter & Cost \\
\midrule
\midrule
Sphere	& $1.28863\times  10^6$  & $8.43504\times  10^2$  & 723  & $8.52218\times  10^2$  & 10000 & $3.28738\times  10^3$ & 1240 & $8.58949\times  10^2$  \\
Torus	& $1.88612\times  10^6$ & $1.21137\times  10^4$   & 847  & $1.23950\times  10^4$  & 6964  & $1.25181\times  10^5$  & 1495 & $1.22184\times  10^4$   \\ 
Parking Garage	& $8.36192\times  10^3$ & $6.31262\times  10^{-1}$   & 117  & $7.93764\times  10^{-1}$ & 5142  & $8.16846\times  10^3$ & 1472 & $6.94700\times  10^{-1}$    \\
Cubicle & $2.53917\times  10^6$ & $3.18310\times  10^2$   & 701  &$3.38483\times  10^2$  & 9709  & $2.20025\times  10^3$ & 244  & $3.97225\times  10^2$   \\ 
Rim	    & $4.06073\times  10^7$  & $1.24992\times  10^3$  & 2355 & $6.50345\times  10^3$   & 6142  & $1.00088\times  10^{23}$ & 2932 & $3.60934\times  10^3$  \\ 
Grid	& $7.21751\times  10^7$  & $4.21596\times  10^4$  & 327  & $4.24620\times  10^4$  & 5613  & $4.96610\times  10^4$  & 1608 & $4.23358\times  10^4$ \\
\bottomrule%
\end{tabular}
\vspace{2mm}
\end{table*}

\subsection{Joining and Leaving the Robot Web}
\label{sec:dynamic}


The Robot Web is fully dynamic because each robot does not need any
information about the group as a whole, so robots can join or leave
freely. 
When new robots are added, randomly into the arena, 
it is initialised at the centre of the arena and starts to
participate in the Robot Web. It does not start to move until it
believes that it has a good pose estimate.
This decision is based on each robot monitoring the
robust scaling of its factors, which is based on the Mahalanobis distance.
Our implementation checks whether  (a) the average scaling for all outgoing factors is $> 0.95$, and (b) that there are at least 8 different observations.
Until these criteria are met, the newly-added robots send empty messages on the inter-robot factors and therefore do not affect the already-initialised robots until they are confident enough to start moving and properly taking part in the web. A video demonstration of this in simulation is available here:\\
\url{https://rmurai.co.uk/projects/RobotWeb#dynamic}


\subsection{Comparison against other solvers}
While the focus of the paper is on distributed localisation using range-bearing sensor measurements, the fact that our method operates on an arbitrary factor graph enables the framework to work with different sensor modalities, for instance, inter-robot \SE3 transformation, often used in distributed pose-graph optimisation (PGO). Here, we solve the following problem:
\beq
\min_{
\substack{\vect_i \in \RR^3, \\ \matR_i \in \SO3, \forall i}
}\frac{1}{2} \sum_{\{i, j\} \in \varepsilon} 
\tau_{ij}  \| \vect_j - \vect_i - \matR_i \vect^z_{i, j} \|^2_2 
+ \kappa_{ij} \| \matR_j - \matR_i \matR^z_{i, j}\|^2_F~,~\label{eq.pgo_problem}
\eeq
where $\varepsilon$ is a set of all measurements, $\matR_i$ is a rotation variable, $\vect_i$ is a translation variable, $\matR^z_{i, j}$ is the measured rotation from $i$ to $j$ and similarly $\vect^z_{i, j}$ is the measured translation from $i$ to $j$.
$\tau_{ij}, \kappa_{ij}$ are the noise parameter computed from the dataset as done in~\cite{Rosen:etal:IJRR, Tian:etal:RAL2020, Murphey:Fan:IROS2020}.

The main complexities of PGO lie in how we handle poor initialisation. As the optimisation problem is non-convex, there exist many local minima. If we directly solve Equation~\eqref{eq.pgo_problem}, we will get stuck in a local minimum, even with small noise~\cite{Carlone:etal:ICRA2015}.
Following~\cite{Choudhary:etal:IJRR2017}, we thus solve a relaxed, linear problem in two stages in a distributed manner.
First, we solve the rotation problem:
\beq
\min_{\matR_i \in \SO3, \forall i} \frac{1}{2} \sum_{\{i, j\} \in \varepsilon} \kappa_{ij}  \| \matR_j - \matR_i \matR^z_{i, j}\|^2_F~.~\label{eq.rotation_problem}
\eeq
We solve the quadratic relaxation of this problem, by dropping the \SO3 constraint and then projecting the solution back to \SO3 via SVD.

We then solve for the full pose using a linear approximation of rotation perturbation:
\bea
\min_{\vect_i, \theta_i\in \RR^3, \forall i}  \frac{1}{2} \sum_{\{i, j\} \in \varepsilon} \tau_{ij}  \| 
\vect_j - \vect_i - \matR_i \tilde\Exp(\theta_i) \vect^z_{i, j}
\|^2_2 \nonumber \\
+ \kappa_{ij}  \| \matR_j\tilde\Exp(\theta_j) - \matR_i \tilde\Exp(\theta_i) \matR^z_{i, j}\|_F^2
~,\label{eq.full_pose_problem}
\eea
where $\tilde\Exp(\theta) = \matI_3 + \matS(\theta)$, and $\matS(\theta)$ is a skew symmetric matrix.

In~\cite{Choudhary:etal:IJRR2017}, Equation~\eqref{eq.rotation_problem}, and~\eqref{eq.full_pose_problem} is solved using distributed Block Gauss-Seidel (DGS) or distributed Block Jacobi method.
Here, we compare GBP and DGS for solving the two-stage PGO problem. We compare against DGS as it is used as an initialisation for other works~\cite{Tian:etal:RAL2020, Tian:etal:TRO2021}, and relaxation is simple to perform with the factor graph framework.
In our evaluation, we report the initial and final cost and the number of iterations required to satisfy the termination condition. We evaluate the trajectories on pose-graph optimisation dataset~\cite{Carlone:etal:ICRA2015}. Each trajectory is split into 50 segments to simulate a multi-robot pose-graph.

The setup of our evaluation favours the DGS. We count one iteration as a full DGS sweep, where the robots sequentially send the updated information to the next robots in a specific order. GBP on the contrary is robot-wise parallel and does not require coordinated updates.
Hence, the communication pattern of GBP is closer to the distributed Block Jacobi method rather than DGS. Furthermore, we enable flagged initialisation for both DGS and distributed Block Jacobi method. 
For all methods, we terminate the iterations once the norm of the change in the rotation or the pose is below a specified threshold. Here, we use $10^{-2}$ as the threshold for both the rotation and the pose update, for all of our distributed solvers as recommended in~\cite{Choudhary:etal:IJRR2017}. Furthermore, for DGS and distributed Block Jacobi, we use the recommended relaxation parameter of 1.0 which we too found to work the best.
We allow all the solvers to run for up to 10000 iterations.

As shown in Table~\ref{table:comparison_dsolvers}, GBP performs comparable to DGS and obtains cost close to the centralised Gauss-Netwon solver, though the setup favours DGS, and DGS has distributed pose-graph specific heuristics such as flagged initialisation. Compared to the distributed Block Jacobi method which has a similar communication pattern as GBP, GBP performs significantly better, both in terms of final cost and the number of iterations.
This result highlights the generality of GBP and makes GBP a promising alternative to the existing distributed solvers.
Devising a fair and complete evaluation of different distributed solvers is an interesting direction for future work.

\section{Demonstrations and Experiments in a Real-World}
\label{sec:realworld}

\begin{figure}[!t]
  \center
  \includegraphics[width=\linewidth]{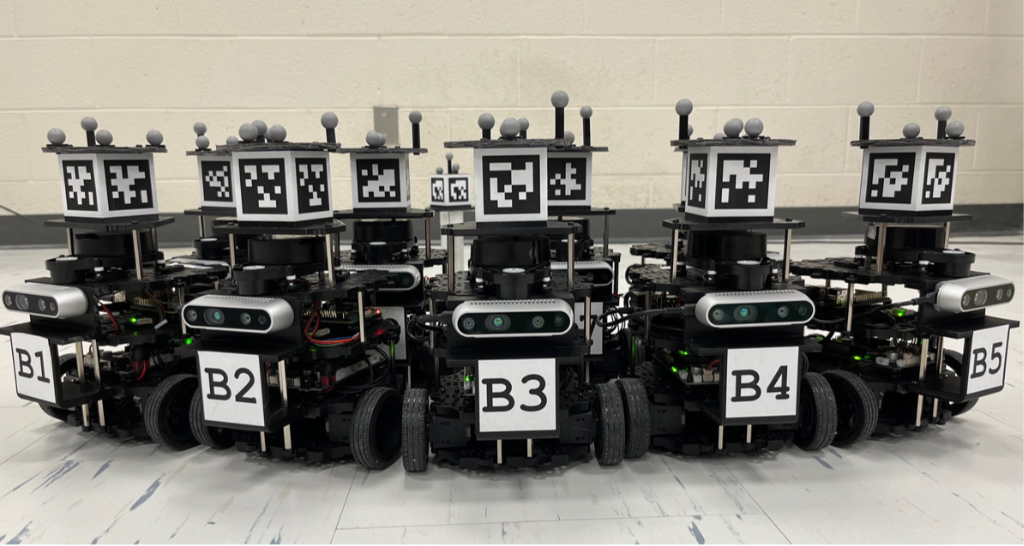}
  \caption{
    Image of a Turtlebot3 Burger used in the real robot experiment. It is fitted with AprilTag-labelled cubes, an Intel-Realsense D435i camera and Vicon markers. Vicon markers are only used to obtain the ground-truth trajectories, used for the evaluation. The depth image, laser scanner and IMU are not used in any of the experiments.
    }\label{fig:robot}
\vspace{2mm} \hrule
\end{figure}

To provide concrete evidence of the effectiveness of our approach, we have evaluated the real robots running our system on onboard devices in a distributed manner.

\subsection{Evaluation Setup}
To evaluate our approach with real robots, we used nine TurtleBot3 Burgers as the robot platform. 
The robots (as shown in Figure~\ref{fig:robot}) were equipped with a Raspberry Pi 3B+ computer with a Cortex-A53 64-bit 1.4GHz processor and 1GB of RAM as the onboard computer. In addition, each robot was fitted with an AprilTag-labeled cube -- with the same tag on all sides -- and an Intel-Realsense D435i camera. The RGB images captured by the camera and the data from the wheel encoder served as the sensory input. To simplify the setup, the depth image, IMU, and laser scanner were disabled, and for the odometry, only the wheel odometry was used.
Each robot had knowledge of the size and location of the AprilTag~\cite{Olson:ICRA2011}, the camera position, and the calibration parameters.
As we have many robots, factory calibration was used for the odometry and the camera. This is unideal as it adds systematic bias; however, our approach was still able to function effectively.

During the described experiment, the robots are instructed to follow a square trajectory. When the Robot Web system detects a drift in the robot's position, a heuristic is used to correct the pose. For the drift of less than 5cm, proportional control is applied to bring the robot back onto the desired trajectory. Otherwise, the robot turns to face the next corner of the trajectory to correct the pose.

All computation, including GBP optimisation, pose correction, and inter-device communication via ROS2, runs on the onboard computer,  highlighting the computational efficiency of our approach. We assume that the robots know the mapping between unique IDs and IP addresses in advance and that there is a shared/synchronised clock for all observations. 
When an AprilTag is detected, observation is transformed into a range-bearing measurement. We are unable to obtain relative transformation measurements -- which include both translation and rotation -- as the same AprilTag is used on all sides of the cube, so the orientation is ambiguous.

\subsection{Implementation Details}

We use ROS2 Foxy~\cite{Macenski:2022} for all the robots. The Publish-subscribe model as described in Section~\ref{sec:comm_model} is used for message passing. In ROS2 this entails simply subscribing to the topics (e.g. for robot 1, it will subscribe to \texttt{robot\_1/variable\_msg}, \texttt{robot\_1/factor\_msg}) and publishing to either the variable/factor of other robots along the inter-device factor.

GBP runs on its own thread, and the internal message passing runs as fast as possible.
GBP is interleaved with the subscriber which receives the inter-device messages. For simplicity, a single coarse lock is used to avoid concurrency problems (adding inter-device messages to the internal factor graph); however, as all update operations of GBP are local, it is possible to use a finer lock. 
The publisher runs at 10Hz, publishing the outgoing messages. Best-effort delivery is used; hence, there is no delivery guarantee.
We emphasise that the publishing and receiving of the messages are not synchronised, and robots receive messages at arbitrary timings (potentially out of order). 

We set the sensor noise to be: $\sigma_x = 0.01$m, $\sigma_y = 0.01$m, and $\sigma_{\theta} = 1 \deg$ for the prior; $\sigma_x = 0.01$m, $\sigma_y = 0.005$m, and $\sigma_{\theta} = 1 \deg$ for the odometry; and $\sigma_b = 0.01$m, and $\sigma_{\theta} = 1 \deg$ for the range-bearing.
All robots run GBP with a window size of 5. DSC~\cite{Agarwal:etal:ICRA2013} is used for the robust kernel with $\Phi = 10$. The variable nodes are after any forward motion or a rotation, and in all the experiments, 75 poses per robot were added to the graph.

\begin{figure*}[!t]
  \begin{subfigure}[b]{0.23\textwidth}
    \includegraphics[width=\textwidth]{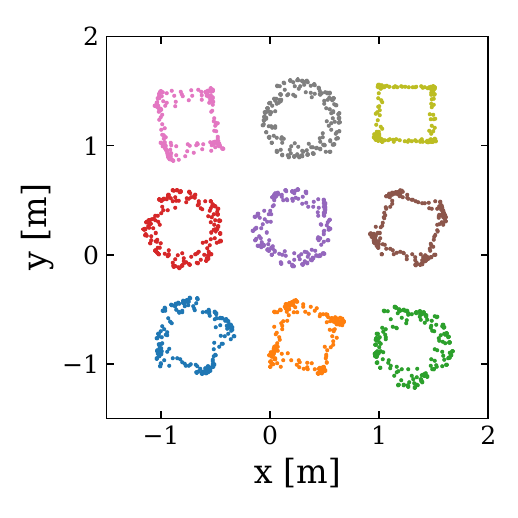}
    \caption{No landmarks, no inter-device communication.}
  \end{subfigure}
  \hfill
  \begin{subfigure}[b]{0.23\textwidth}
    \includegraphics[width=\textwidth]{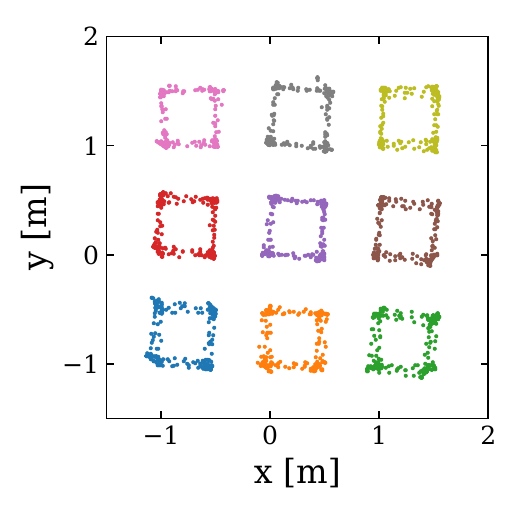}
    \caption{No landmarks, with inter-device communication.}
  \end{subfigure}
  \hfill
  \begin{subfigure}[b]{0.23\textwidth}
    \includegraphics[width=\textwidth]{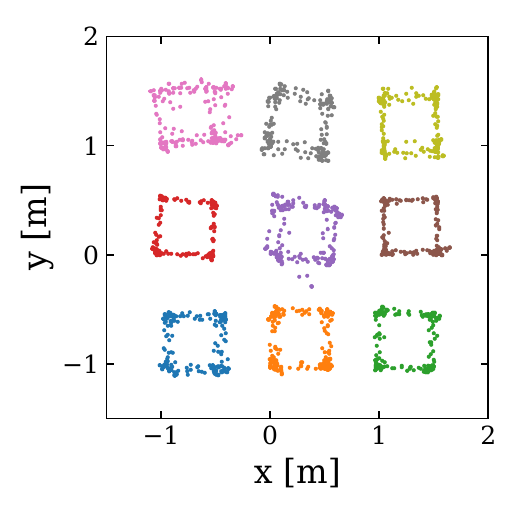}
    \caption{With landmarks, no inter-device communication.}
  \end{subfigure}
  \hfill
  \begin{subfigure}[b]{0.23\textwidth}
    \includegraphics[width=\textwidth]{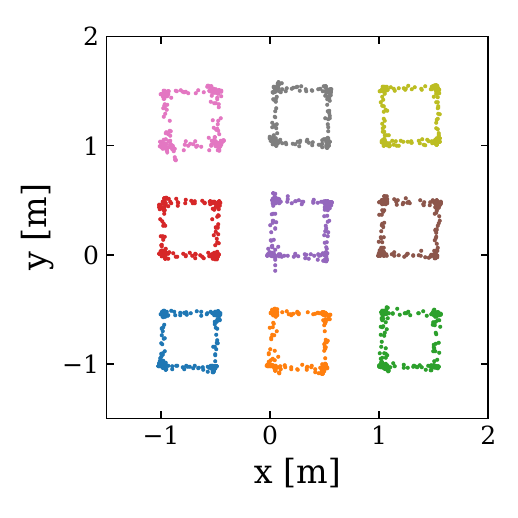}
    \caption{With landmarks, with inter-device communication.}
  \end{subfigure}
  \hfill
  \caption{
  Nine real robots are moving in a square trajectory, and the motion captured by the Vicon system is plotted. It is clearly visible that using inter-device communication improves the localisation accuracy.
  }
\label{fig:real_robot_traj}
\vspace{2mm} \hrule
\end{figure*}

\begin{table}[!t]
\caption{
\label{table:vicon_cmp}
The table below shows the RMSE ATE of the real robot experiment. The RMSE ATE of the real robots is computed against the observations made by the Vicon motion capture system.
The table summarises the impact of inter-device communication and the availability of landmarks on the RMSE ATE. We report the mean error and the standard deviation of the nine robots.
}
\centering
\begin{tabular}{ccc}
\toprule

Communication & Landmark & $\mu\pm\sigma$ [m] \\
\midrule
\midrule
False & False & $0.162\pm0.085$  \\
True & False  & $\mathbf{0.043\pm0.020}$  \\
\midrule
False & True  & $0.071\pm0.020$  \\
True & True   & $\mathbf{0.028\pm0.007}$  \\
\bottomrule%

\end{tabular}
\vspace{2mm}
\end{table}

\subsection{Multi-Robot Localisation Evaluation}
In this section, we evaluate the localisation accuracy of our approach.
We evaluate under two different settings, with and without landmarks.
Four landmarks are used, and their position is known to the robots in advance.
Figure~\ref{fig:real_robot_traj} shows the trajectory captured by the Vicon motion capture system. In all runs, robots are moving for 10 minutes.
It is clear that Robot Web localises the robots well and allows them to operate for a long period without drifting.

The RMSE ATE of the real robots is computed against the observations made by the Vicon motion capture system. The result is summarised in Table~\ref{table:vicon_cmp}. As expected whether there are landmarks or not, using inter-device communication, i.e. using Robot Web, improves the localisation accuracy. 
The use of sparse landmarks is insufficient for good localisation without inter-device communication. This is clear both qualitatively by comparing (b) and (c) of Figure~\ref{fig:real_robot_traj}, and quantitatively in Table~\ref{table:vicon_cmp} by comparing: no landmarks, with inter-device communication; and with landmarks, no inter-device communication.

\subsection{Relocalisation Demonstration}
\begin{figure*}[!t]
  \begin{subfigure}[b]{0.32\textwidth}
    \includegraphics[trim={300 0 300 0},clip,width=\textwidth]{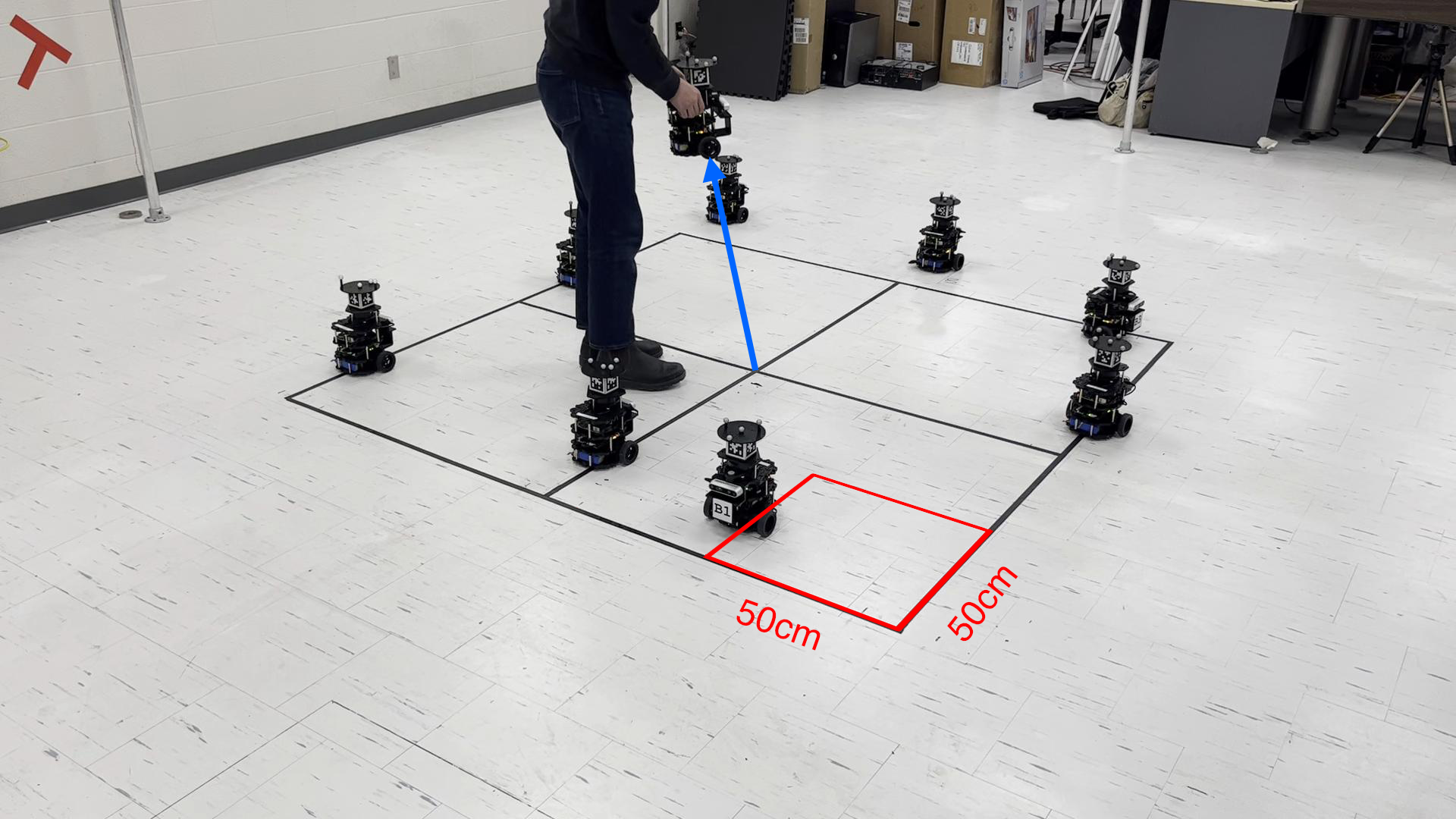}
    \caption{}
  \end{subfigure}
  \hfill
  \begin{subfigure}[b]{0.32\textwidth}
    \includegraphics[trim={300 0 300 0},clip,width=\textwidth]{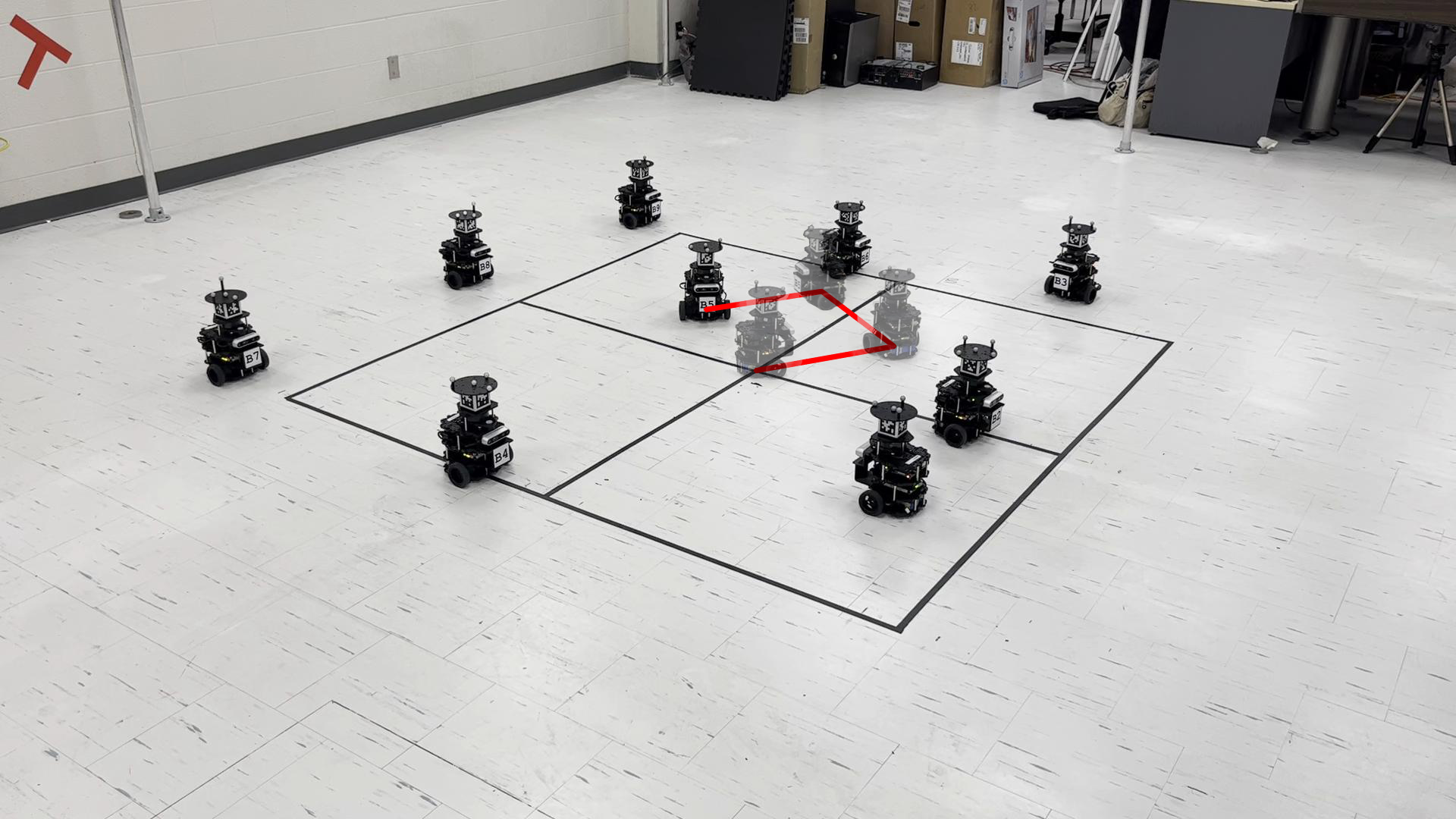}
    \caption{}
  \end{subfigure}
  \hfill
  \begin{subfigure}[b]{0.32\textwidth}
    \includegraphics[trim={300 0 300 0},clip,width=\textwidth]{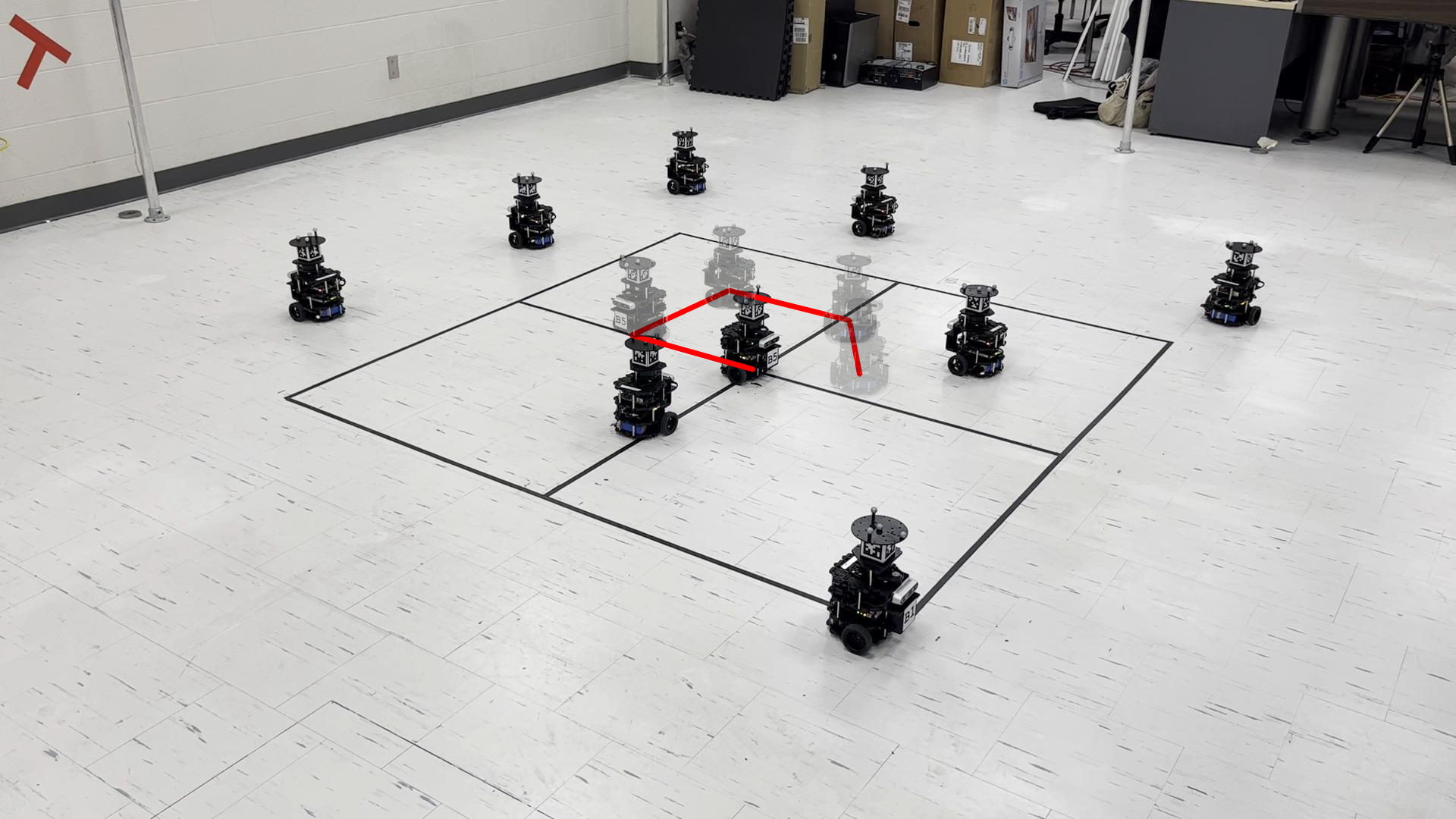}
    \caption{}
  \end{subfigure}
  \hfill
  \caption{
    Here, nine robots are running Robot Web. Each robot starts on the vertices of the grid on the floor and moves in a square pattern (50cm x 50cm).
    In (a) during operation, one robot is removed from the system (e.g. for maintenance) and then added back with an incorrect pose. As a result, the robot fails to follow the square pattern, as shown in (b).
    However, using Robot Web, the robot is able to successfully relocalise, as shown in (c), and returns to following the square trajectory. 
  }
\label{fig:robot_web_reloc}
\vspace{2mm} \hrule
\end{figure*}

In a multi-robot system, there are many potential sources of failure for the robots. For instance, a robot might need to be stopped for maintenance due to a low battery, or it could be accidentally bumped out of position by a person. These types of external influences are often non-Gaussian, and if the system only accounts for Gaussian noise, it will not be able to accurately handle these unexpected events.

In Robot Web, while GBP assumes a Gaussian noise, robust factors allow the system to handle non-Gaussian noise as well. In Figure~\ref{fig:robot_web_reloc}, we lift a moving robot and place it back in the wrong position. This disorients the robot, and it is unable to follow the square trajectory. However, after a few observations, the robot relocalises and returns to follow the square trajectory.
During this relocalisation process, other robots are unaffected by the wrongly positioned robot as the robust factor heavily down weights its influence until the wrongly positioned robot is correctly localised.
Due to the error between the position of the robot and its estimate, the measurements made of this robot by the others will have high residuals, and thus will be down-weighted by the robust kernel.
A video of the relocalisation demo is available here: \\\url{https://rmurai.co.uk/projects/RobotWeb#reloc}

\section{Ongoing Research Topics}
\label{sec:ongoing}

We have demonstrated the essential operation of the Robot Web both in a simulation and in a real-world, truly distributed implementation on multiple robots.
The properties of the method are extremely promising, and here we discuss some important research directions going forward.


\subsection{More General Parameterisation}
\label{sec:genparam}

Our current implementation makes several simplifying assumptions, but
we believe that all of these are fairly straightforward to remove
within the Robot Web framework with some further work.

\bi
\item We currently assume that inter-robot measurement factors, stored
  by the robot with the sensor, always correspond to observations of
  the position of the centre of the second robot. This would already allow a practical implementation for 2D planar robots which each carry a single observable beacon above their  centres.
  More realistically,
  each robot might  have several or many observable features, and
  these will be located on any point on its structure. We can deal
  with this by adding additional internal variables to the second
  robot, connected to its main pose by `perfect' factors, representing
  the positions of the observable features relative to its body, with
  positions that only need to be known to the second robot.
\item
Our current assumption that all robots have pose variables defined at the same
rate and at corresponding times could be relaxed by measurement
factors which connect to multiple variables at  the receiving robot and interpolate the measurement between poses.
\ei

We might take the Robot Web idea even further to also apply {\em
  inside} a single robot's modular body. The different parts,
actuators and sensors that make up the robot might use Web interfaces
between them to enable distributed joint estimation and very general modularity.

\subsection{Efficient Long-Term Operation}
\label{sec:long}

If we keep the full history of all pose variables for each robot, and
all measurement factors, eventually the computation, storage and
communication capacity of each robot  would become overloaded. Of
these,  inter-robot communication  is likely to be the  main
bottleneck. 
We showed one simple approach
to dealing  with this  via time windowing, where poses older than a
threshold are discarded, and this gives good performance when robots
have bounded drift due to the presence of known  beacons.

A more
general approach to bounding  the growth  of  the graph could be based
on  incremental abstraction~\cite{Ortiz:etal:ICRA2022}, where  past variables and factors are not
deleted but grouped into more efficient  blocks with   minimal loss of
accuracy. For  instance, a set of well-estimated  pose variables from the  past
could be grouped into  an abstract trajectory  segment, represented by
far  fewer variables. Factors could  also be grouped. Achieving this
incremental abstraction in a fully distributed way across multiple
robots however will require  substantial  research.

\section{Discussion and Conclusions}
\label{sec:conclusion}

We have presented a method for distributed multi-robot
localisation in the context of a larger `Robot Web' vision for how
heterogeneous groups of intelligent robots and devices of the future
could cooperate and coordinate.
This approach could be important at a time when many different
companies and organisations are building spatially aware devices, and
offers a distributed, inter-operable alternative to a single unified cloud maps solution.

As the performance and scale of many-robot systems may greatly improve
due to work such as ours, it is important to consider potential
ethical concerns. A robust, large-scale robot group or `swarm' has
many possible positive applications, such  as the automation of
farming or environmental surveillance via many low-cost  devices,
which could be much more efficient overall than a small number of
large devices. 
However, there are possible ethical concerns with swarms of autonomous, weaponised drones


We believe that our paper overall could indicate a positive direction for
the operation of distributed multi-robot systems via the specification
that the Robot Web allows and demands of an {\bf open communication protocol}.
If the majority of the moving intelligent devices were to take part in
such a system by publishing and reading localisation messages via this
open protocol, it would be greatly to  the advantage of any newly
built devices to also take part, to exchange open messages, and to benefit from the system. This
would mean that the whole system might work in a way similar to the
World Wide Web, and some degree of global control would be possible via the interpretability of the protocol and perhaps more specific 
safety measures built into it. We believe that it is better
for devices to be exchanging clearly interpretable geometric information
than cryptic coded messages (as would emerge for instance in a possible distributed `graph neural network' system for localisation, where the format of messages is learned rather than designed --- and we should add here our view that a learned alternative to our method is also likely to be far less flexible and efficient).

These are ongoing issues to be debated as the technology advances,
and we as authors believe that researchers
should openly engage with these issues and play a part in designing
the correct principles into the technology.

In the longer term future, the distributed coordination of intelligent
moving systems is a key part of the concept of `intelligent matter',
where distribution and communication might be at the microscopic
level to enable new classes of technology such as micromachines~\cite{Huang:etal:AR2021} which can self-organise in
ways that might approach the capabilities of biological systems
\cite{Kaspar:etal:Nature2021}. 
Recently, it has been shown that essentially the same computation framework that we have demonstrated in Robot Web using GBP can also be applied to multi-robot motion planning~\cite{aalok:etal:RAL2023}.
Efficient, robust distributed localisation will be
one of the most important enabling layers of such systems.

\section*{Acknowledgements}
We are grateful to many researchers with whom we have discussed some of the ideas in this paper, especially from the Dyson Robotics Lab and Robot Vision Group at Imperial College, and SLAMcore. We would particularly like to thank Aalok Patwardhan, Marwan Taher, Hussein Ali Jaafar, Stefan Leutenegger, Raluca Scona, Callum Rhodes, Ignacio Alzugaray, Talfan Evans, Eric Dexheimer, Seth Nabarro, Mark van der Wilk, Owen  Nicholson, Rob Deaves, Pablo Alcantarilla, Jacek Zienkiewicz, Amanda Prorok, Mac Schwager, Frank Dellaert, Michael Kaess, Tim Barfoot, Richard Newcombe.
This work was partially supported by the EPSRC (EP/K008730/1 and EP/P010040/1).

\bibliographystyle{plain} 
\bibliography{robotvision}

\end{document}